\documentclass[10pt,twocolumn,letterpaper]{article}
\usepackage{booktabs}
\usepackage{multirow} 
\usepackage[table]{xcolor}
\usepackage{amsthm}
\usepackage{amssymb}
 
\usepackage{array}
\usepackage[pagenumbers]{cvpr} 

\usepackage{lineno}
\definecolor{cvprblue}{rgb}{0.21,0.49,0.74}
\usepackage[pagebackref,breaklinks,colorlinks,allcolors=cvprblue]{hyperref}

\def\paperID{30825} 
\def\confName{CVPR}
\def\confYear{2026}

\title{Dr. Seg: Revisiting GRPO Training for Visual Large Language Models through Perception-Oriented Design}

\makeatletter
\renewcommand{\@fnsymbol}[1]{\ensuremath{\dagger}}
\makeatother

\author{
Haoxiang Sun\textsuperscript{1}\quad
Tao Wang\textsuperscript{1}\thanks{Corresponding author. }\quad
Chenwei Tang\textsuperscript{1}\quad
Li Yuan\textsuperscript{2}\quad
Jiancheng Lv\textsuperscript{1}\\
\textsuperscript{1}College of Computer Science, Sichuan University\\
\textsuperscript{2}School of Electronic and Computer Engineering, Peking University\\
{\tt\small haoxiang0311@gmail.com, twangnh@gmail.com}
}

\begin{document}
\maketitle
\begin{abstract}
Following the success of Group Relative Policy Optimization (GRPO) in foundation LLMs, an increasing number of works have sought to adapt GRPO to Visual Large Language Models (VLLMs) for visual perception tasks (e.g., detection and segmentation). 
However, much of this line of research rests on a long-standing yet unexamined assumption: training paradigms developed for language reasoning can be transferred seamlessly to visual perception.
Our experiments show that this assumption is not valid, revealing intrinsic differences between reasoning-oriented and perception-oriented settings. Using reasoning segmentation as a representative case, we surface two overlooked factors: (i) the need for a broader output space, and (ii) the importance of fine-grained, stable rewards. 
Building on these observations, we propose Dr.~Seg, a simple, plug-and-play GRPO-based framework consisting of a Look-to-Confirm mechanism and a Distribution-Ranked Reward module, requiring no architectural modifications and integrating seamlessly with existing GRPO-based VLLMs.
Extensive experiments demonstrate that Dr.~Seg improves performance in complex visual scenarios while maintaining strong generalization. Code,  models, and datasets are available at \url{https://github.com/eVI-group-SCU/Dr-Seg}.

\end{abstract}    
\vspace{-10pt}
\section{Introduction}
\label{sec:intro}

\begin{figure}[t]
  \centering   \includegraphics[width=1.0\linewidth]{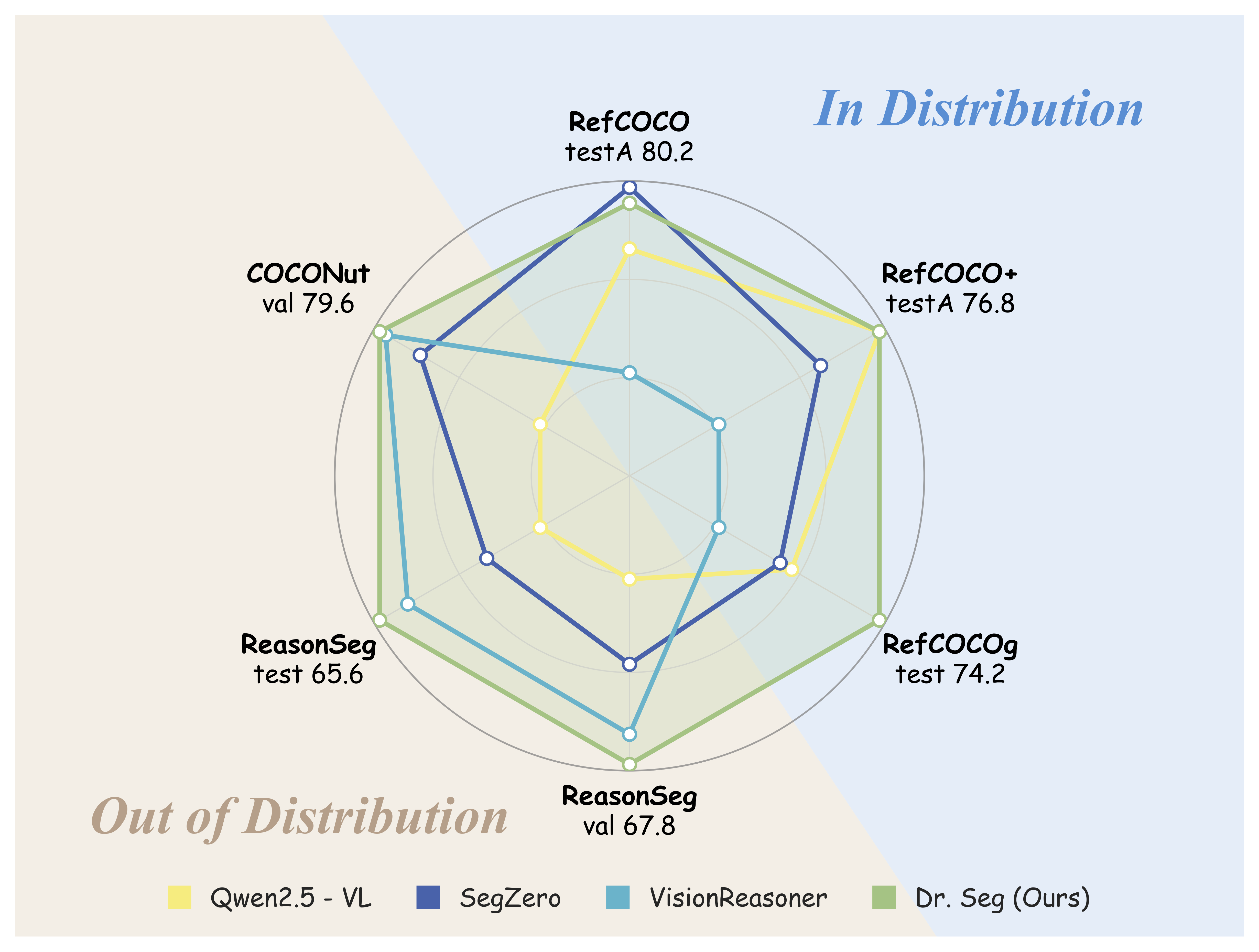}
   \vspace{-20pt}
   \caption{Dr.~Seg achieves new state-of-the-art results on 5 out of 6 benchmarks under both in-distribution (ID) and out-of-distribution (OOD) conditions, demonstrating strong generalization ability.}
   \label{fig:onecol}
   \vspace{-12pt}
\end{figure}


Visual segmentation is a fundamental task in computer vision and has traditionally been dominated by specialist models~\cite{yang2022lavt,ding2021vision,liu2023gres}. In recent years, Visual Large Language Models (VLLMs) have demonstrated strong visual understanding capabilities through large-scale vision-language pretraining~\cite{bai2025qwen2,liu2023visual,hurst2024gpt,chen2024internvl}. By further applying instruction tuning on specially curated datasets, VLLMs have extended their capabilities to more complex and fine-grained perception tasks~\cite{zhang2024gpt4roi,li2024groundinggpt,ranasinghe2024learning}. 

Early attempts have further extended the capabilities of VLLMs towards reasoning-oriented segmentation, enhancing their flexibility and practical applicability~\cite{lai2024lisa,zhang2024psalm,zhang2024omg}. Despite these advances, models that rely solely on instruction tuning exhibit limited generalization and are prone to catastrophic forgetting of general capabilities~\cite{liu2025seg}.
Recent studies, inspired by DeepSeek’s Group Relative Policy Optimization (GRPO)~\cite{shao2024deepseekmath} method, have begun to adapt Reinforcement
Learning with Verifiable Rewards (RLVR) to VLLMs in the post-training stage for a variety of vision tasks~\cite{liu2025seg,shen2025vlm,liu2025visionreasoner}. Despite differences in motivation and application domains, these methods generally follow a common pattern: they conduct post-training on VLLMs, curate task-specific datasets, and design reward functions to achieve performance improvements on targeted tasks. These efforts have led to substantial gains in the generalization of vision tasks, opening up new directions.

However, we find that directly transferring training strategies from reasoning-oriented tasks to perception-oriented tasks is not optimal. The suboptimality is rooted in fundamental differences in learning dynamics and reward design: (i) perception-oriented tasks inherently require a broader output space to accommodate more diverse visual information, whereas reasoning-oriented tasks are driven by causal chains that naturally induce depth-oriented exploration within a narrower output space; and (ii) perception-oriented tasks demand fairer, better-calibrated rewards to reduce interference and noise among multiple optimization objectives. 
We thus aim to develop more effective learning methods that encourage broader output space exploration and provide finer reward feedback.


 

To this end, we introduce Dr.~Seg, a simple yet effective GRPO-based framework for VLLMs, specifically tailored for visual perception tasks. Dr.~Seg is built upon two key components.
First, it integrates a Look-to-Confirm strategy that explicitly broadens the output search space: the model is encouraged to show its focus on certain visual information during reasoning, thus it is forced to look for plausible visual cues before confirming its final decision. This design promotes breadth-oriented exploration by enabling the model to derive reasoning paths from diverse dimensions (e.g., shape, material, relations) within its pre-trained visual knowledge, and in turn yields stronger generalization. 
Second, it incorporates a Distribution-Ranked Reward mechanism that provides fine-grained and adaptive feedback by tracking performance variations across training steps, while being robust to heterogeneity in metric scales, thus guiding the model toward more precise and stable improvements. \textit{Together, these two components reinforce each other, with finer-grained reward to learn more accurate prediction under the wider output exploration space, yielding gains that go beyond the sum of their individual contributions}. 
Notably, Dr.~Seg requires no modifications to the underlying model architecture, making it both lightweight and broadly applicable. 
It outperforms prior work on core vision tasks including referring expression comprehension and segmentation, object detection, reasoning segmentation and counting, e.g., compared to the baseline method, Dr.~Seg achieves improvements of 2.0 absolute gIoU for ReasonSeg-test segmentation benchmark, 2.4 absolute AP for COCO detection benchmark and 4.5 for Pixmo-val counting benchmark.

In summary, our contributions are as follows:
\begin{itemize}
    \item We identify key differences in GRPO training for visual perception tasks and reasoning tasks, i.e., output space and reward designs, and present a detailed analysis.
    \item We design a Look-to-Confirm  strategy and Distribution-Ranked Reward mechanism, which are integrated to form a plug-and-play Dr.~Seg method and collaboratively enhance the learning of visual perception in VLLMs.
    \item We present a challenging COCONut dataset to evaluate multi-object perception. Extensive experiments show that Dr.~Seg achieves superior performance on multiple perception tasks while preserving robust generalization.

\end{itemize}

\section{Related Work}
\vspace{-6pt}
\textbf{Reasoning Segmentation with VLLMs.} The emergence of Visual Large Language Models has significantly advanced visual understanding by combining perception with their inherent reasoning capabilities~\cite{bai2025qwen2,chen2024internvl,hurst2024gpt}. Many existing works leverage the next-token prediction paradigm by generating special tokens to guide a mask decoder for segmentation, inevitably requiring supervised fine-tuning (SFT)~\cite{lai2024lisa,zhang2024omg,xia2024gsva,rasheed2024glamm,ren2024pixellm}. Another line of research generates prompts (e.g., boxes or points) to guide SAM~\cite{kirillov2023segment}, thereby offering greater flexibility~\cite{liu2025seg,liu2025visionreasoner,wang2025pixelthink}. Our work builds upon the latter approach, aiming to preserve both simplicity and flexibility.

\noindent\textbf{Reinforcement Learning in VLLMs.} Following the success of DeepSeek-R1~\cite{guo2025deepseek}, a series of works have explored reinforcement learning methods in multimodal domains. Perception-R1~\cite{yu2025perception} and VisionReasoner~\cite{liu2025visionreasoner} address multiple perception-oriented tasks and apply GRPO during post-training. In parallel, Seg-Zero~\cite{liu2025seg}, VLM-R1~\cite{shen2025vlm}, VersaVid-R1~\cite{chen2025versavid}, and CrowdVLM-R1~\cite{wang2025crowdvlm} focus on their respective domains. In this work, we focus on the task of Reasoning Segmentation, which serves as a representative setting that simultaneously requires reasoning ability to identify the correct target and perceptual precision to perform accurate segmentation.

\noindent\textbf{Entropy and Output Space Dynamics in RL. }
Recently, a line of studies has begun to investigate where the performance gains in RL truly originate. Chu et al.~\cite{chu2025sft} argue that SFT primarily enables memorization, while RL serves to generalize this ability. Liu et al.~\cite{liu2025prorl} and Wu et al.~\cite{wu2025invisible} both investigate the capability boundaries of RL, yet arrive at divergent conclusions regarding whether RL can transcend the capability space established during pre-training. Another line of work has explored the entropy perspective. Cui et al.~\cite{cui2025entropy} analyze entropy dynamics during training and propose an empirical law, while Wang et al.~\cite{wang2025beyond} reveal that high-entropy tokens play a significant role in the learning process. However, these studies are limited to reasoning-oriented scenarios, whereas our work is, to the best of our knowledge, the first to examine entropy dynamics from the perspective of perception-oriented tasks.

\setlength{\tabcolsep}{3.5pt}{


\begin{figure}[]
    \centering
    \includegraphics[width=1\linewidth]{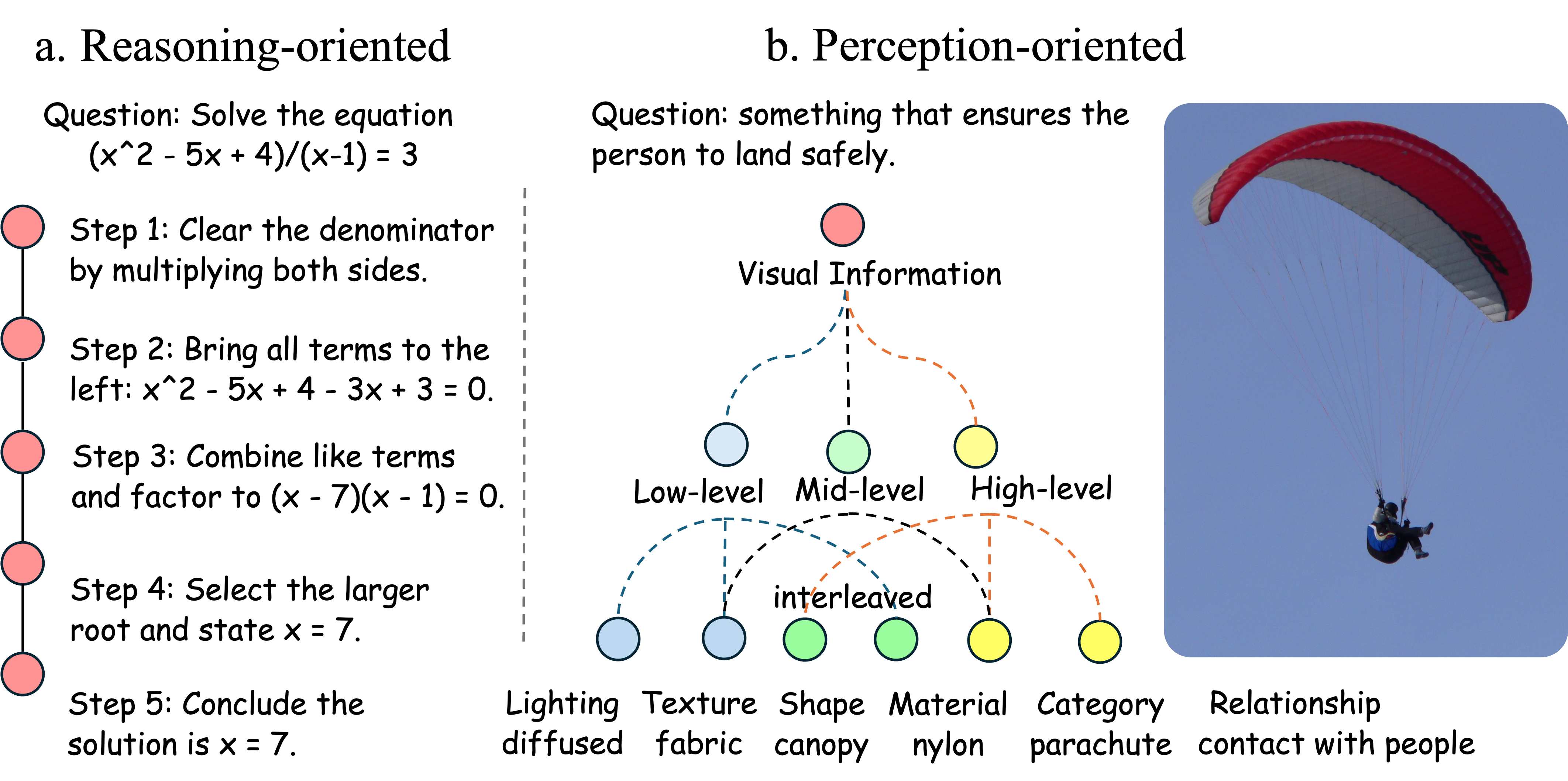}
    \caption{Illustration of breadth-oriented exploration in perception-oriented tasks. Multiple level visual attributes can support different reasoning trajectories.}
    \label{breadth illustration}
    \vspace{-6pt}
\end{figure}

\vspace{-6pt}
\section{Revisiting GRPO for Visual Perception}
\vspace{-6pt}
In this section, we present our findings on the GRPO training process of visual large language models, identifying the differences and challenges w.r.t. reasoning-oriented tasks.

\vspace{-6pt}
\subsection{Preliminaries}
\vspace{-6pt}

\noindent\textbf{Group Relative Policy Optimization.}
Given a query $q$, the policy model $\pi_{\theta_{\text{old}}}$ samples a group of candidate outputs ${o_1, o_2, \dots, o_G}$ for each input.
The policy is then optimized by maximizing the following objective:  
{
\begin{equation}
\begin{aligned}
J_{\mathrm{GRPO}}(\theta) 
&= \mathbb{E}_{q \sim P(Q),\, \{o_i\}_{i=1}^G \sim \pi_{\theta_{\mathrm{old}}}(O \mid q)} \\
&\Biggl[
  \frac{1}{G} \sum_{i=1}^G
  \min\!\Bigl(
    s_1 A_i, s_2 
    A_i
  \Bigr) -\,\beta\,D_{\mathrm{KL}}\bigl(\pi_{\theta}\,\big\|\,\pi_{\mathrm{ref}}\bigr)
\Biggr],\\
s_1 & = \frac{\pi_\theta(o_i|q)}{\pi_{\theta_{old}}(o_i|q)}, \quad s_2  = \text{clip}\left(s_1,1-\epsilon,1+\epsilon\right).
\end{aligned}
\label{training objective}
\end{equation}
}

\noindent where $\epsilon$ and $\beta$ are the PPO clipping hyper-parameter and the coefficient controlling the KL penalty~\cite{schulman2017proximal}. $A_i$ is the computed advantage using the group rewards $\{r_1,r_2,\cdots,r_G\}$, normalized within each group.
{
\begin{equation}
\begin{aligned}
A_i = \frac{r_i - \operatorname{mean}(\{r_1, r_2, \dots, r_G\})}
           {\operatorname{std}(\{r_1, r_2, \dots, r_G\})}
\end{aligned}
\label{eq2}
\end{equation}
}

\noindent\textbf{Reward Composition.}
In our settings, for each candidate output $o_i$, the total reward is composed of two parts: 
a format reward $r_{\mathrm{fmt}}(o_i)$, which evaluates whether the output 
conforms to the required structure, and an accuracy reward $r_{\mathrm{acc}}(o_i)$, 
which measures the alignment between the predicted boxes/points and the ground-truth. 
The final reward is defined as the sum of the two:
\begin{equation}
r_i = r_{\mathrm{fmt}}(o_i) \;+\; 
r_{\mathrm{acc}}(o_i),
\label{eq:reward}
\end{equation}

\noindent
\subsection{Exploration Path and Entropy Fluctuations}
\label{Exploration Path and Entropy Fluctuations}
Unlike reasoning-oriented tasks~\cite{cui2025entropy,wang2025beyond}, we assume that visual perception tasks naturally favor breadth-first exploration.
Specifically, in reasoning-oriented settings such as mathematics and other scientific domains, the conclusions are tightly constrained by the preceding premises. Such a structure naturally favors depth-oriented exploration and tends to narrow the output search space. By contrast, perception-oriented tasks do not fundamentally rely on such rigid ordering and are inherently characterized by a much more diverse information landscape. 
For a given answer, the model can follow many different reasoning trajectories: 
it may ground its decision in low-level cues such as lighting and texture, in mid-level visual attributes such as shape, color, and material, or in higher-level semantic information such as object categories and their spatial relations. 
And there could be combinations between different levels of information.
Many of these diverse reasoning paths can ultimately lead to the correct answer. We provide an example in Fig.~\ref{breadth illustration}. 



We further examine the model’s entropy to corroborate this hypothesis. Following prior work~\cite{cui2025entropy,wang2025beyond}, we plot the training-time token-level average entropy and find that it remains highly unstable, which indicates the model is uncertain for its token outputs during the training, resulting in broader exploration paths.
This is in sharp contrast to recent RLVR reports on reasoning-oriented settings that document a smooth, monotonic descent~\cite{cui2025entropy,wang2025beyond}, as shown in Fig.~\ref{fig:training entropy}. 
We find that encouraging such breadth exploration in the reasoning process can improve performance (66.1 vs. 65.5 in ReasonSeg task), with the entropy fluctuating even more substantially, i.e., the yellow curve in Fig.~\ref{fig:training entropy}. 
Furthermore, the embeddings are markedly more dispersed with such encouragement, as shown in Fig.~\ref{pca}.
The detailed breadth-encouraging strategy named Look-to-Confirm is presented in the next section.
This behavior indicates that, as training progresses, the policy does not collapse into a narrow region of the output space during its exploration.

\begin{figure}
  \centering
  \includegraphics[width=1.0\linewidth]{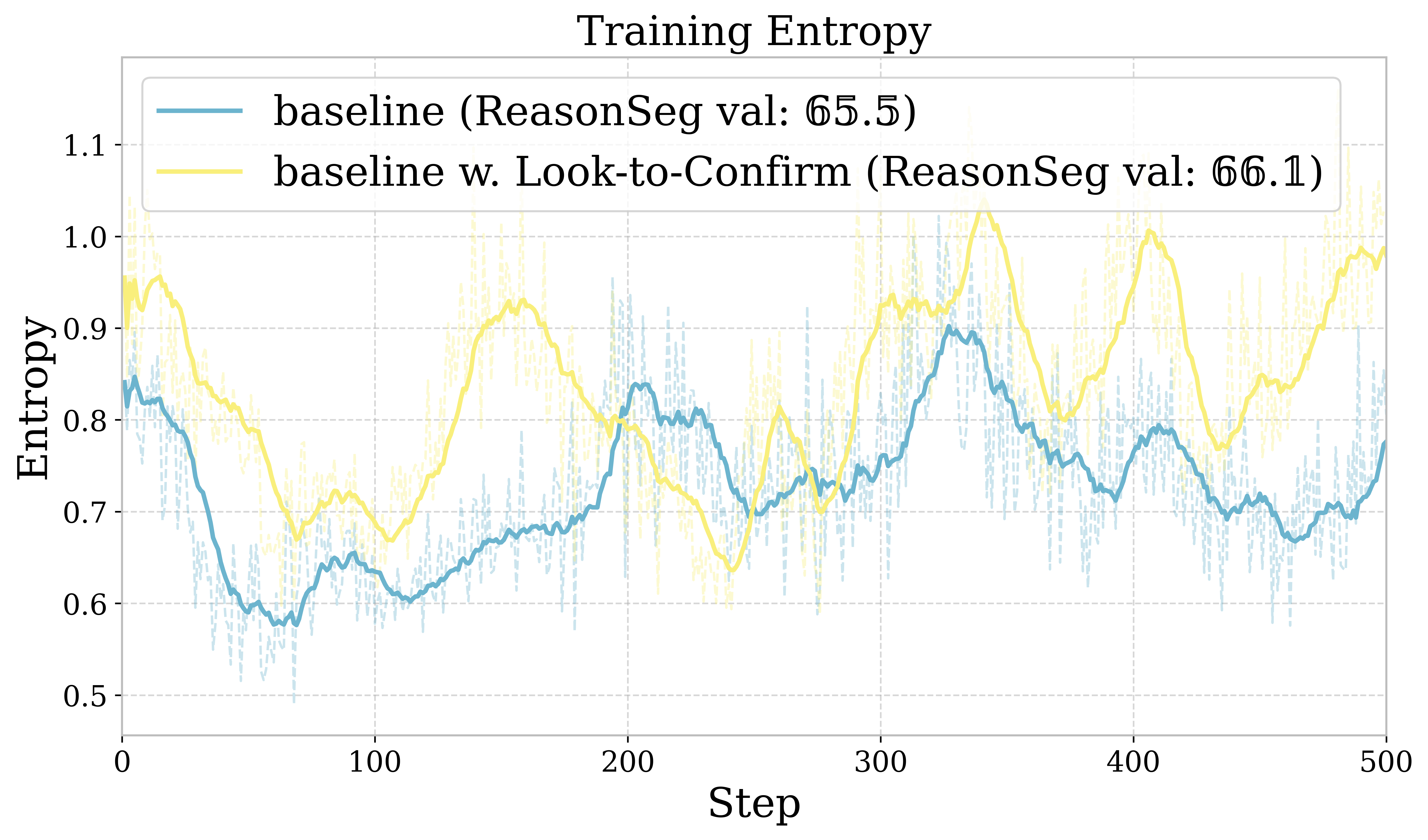}

   \caption{Comparison of token-level entropy during training between baseline VisionReasoner\cite{liu2025visionreasoner} model and baseline with a Look-to-Confirm strategy introduced in the next section. The numbers in parentheses are the performance on ReasonSeg dataset.}
   \label{fig:training entropy}
   \vspace{-12pt}
\end{figure}
\begin{figure}[h]
    \centering
    \includegraphics[width=1\linewidth]{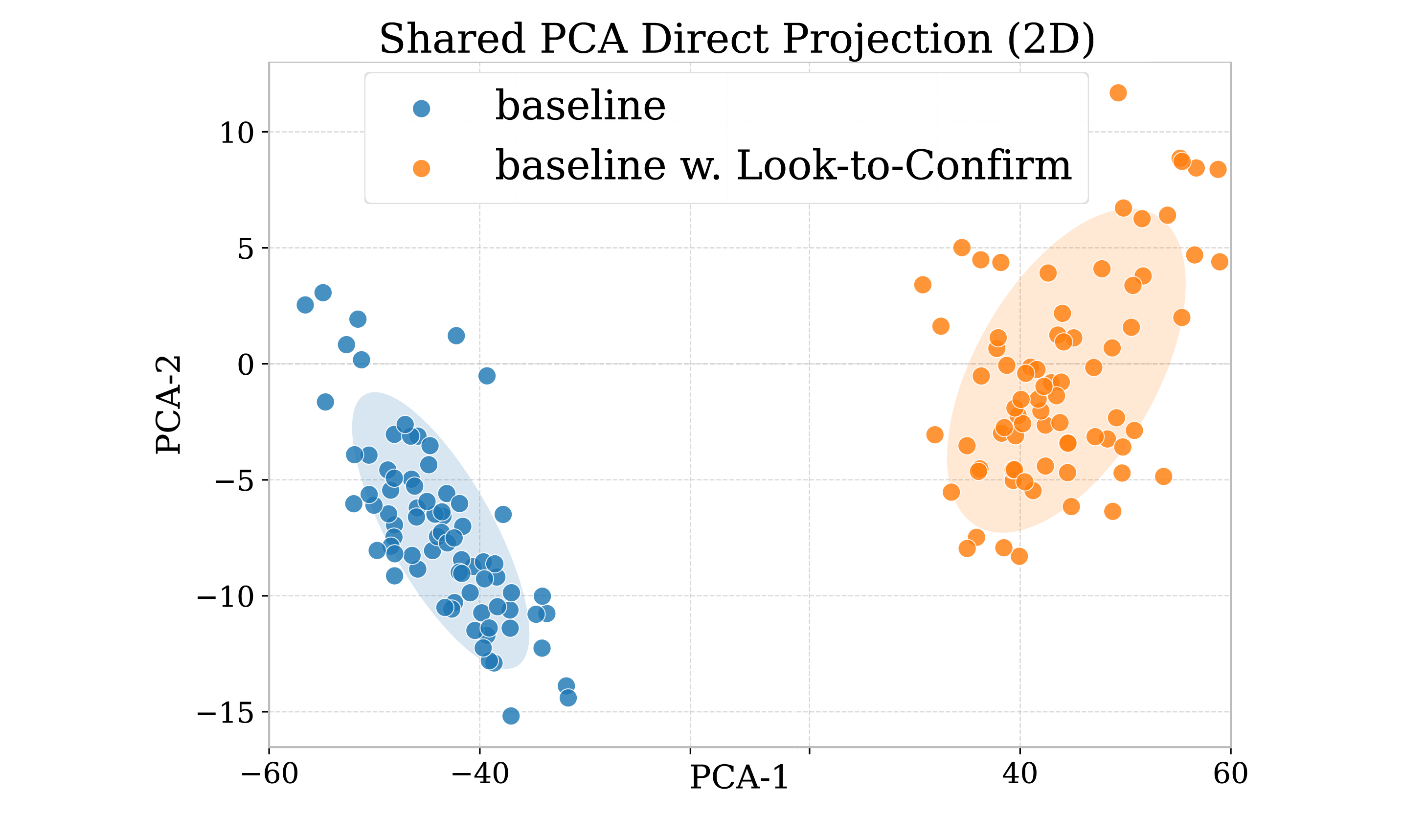}
    \caption{PCA visualization of last-token embeddings for baseline VisionReasoner\cite{liu2025visionreasoner} model vs. baseline with Look-to-Confirm. Points are 2D projections; shaded ellipses denote $2\sigma$ covariance contours around each group mean.}
    \label{pca}
    \vspace{-6pt}
\end{figure}

\noindent
\subsection{The Pitfall in Reward Design}

In addition to the above analysis of the exploration path, the reward design also requires detailed inspection.
Most prior work on VLLM-based perception adopts the binary rewards popular in reasoning tasks~\cite{liu2025visionreasoner,liu2025seg,wang2025pixelthink}. This is counterintuitive for vision, where metrics such as IoU are inherently continuous and thus enable finer reward feedback; collapsing them to binary signals appears unnecessarily lossy. 
A natural next step is to apply raw, unnormalized continuous rewards\footnote{More experiment details in supplementary materials.}. However, as shown in Fig.~\ref{tab:look}, we find that when multiple optimization targets are involved simultaneously, simply using the final reward as the sum of $N$ different accuracy rewards introduces bias, which in turn causes policy optimization to suffer from suboptimality. 

\begin{table}[]
  \caption{Evaluation metrics of the baseline and the raw reward setting. Although the model achieves improved performance on the simpler RefCOCO, RefCOCO+, and RefCOCOg benchmarks, its performance degrades on the more challenging ReasonSeg benchmark, which features a mixture of single- and multi-object cases.}

  \label{tab:look}
  \centering
  \begin{tabular}{lcccccc}
    \toprule
    Method & RCO & RCO+ & RCOg & ReasonSeg \\
    \midrule
    Baseline &  79.0  & 75.3 & 72.5 & \textbf{65.5}\\
    Raw reward  & \textbf{80.0} & \textbf{76.3} & \textbf{74.0 }& 64.7 \\
    \bottomrule
  \end{tabular}
  \vspace{-10pt}
\end{table}

We provide a mathematical explanation showing that the gradient is sensitive to the intrinsic scales of individual reward components $r^{(j)}$; high-variance components dominate the update while low-variance components are suppressed. We assume component-wise finite second moments for the reward vector.
%
For a given query $q$ and output $o$, let the $N$-dimensional reward
vector be $\mathbf{r} := (r^{(1)},\dots,r^{(N)})^\top$ with mean vector
$\boldsymbol{\mu} = (\mu^{(1)},\dots,\mu^{(N)})^\top$ and covariance matrix
$\Sigma = [\sigma^{(jk)}]_{j,k=1}^N$, where $\sigma^{(jk)} = 
\operatorname{Cov}(r^{(j)}, r^{(k)})$. The total scalar reward is
\begin{align}
  r = \sum_{j=1}^N r^{(j)}
\end{align}

Let $\mu_r := \mathbb{E}[r]$ and 
$\sigma_{\mathrm{mix}}^2 := \operatorname{Var}(r) =
\mathbf{1}^\top \Sigma \mathbf{1}$ denote the mixed mean and variance of all $N$ reward components into
a single scale. 
Under the standard GRPO normalization $A = (r - \mu_r) / \sigma_{\mathrm{mix}}$, 
omitting clipping, KL penalties, and importance ratio terms in a simplified form, the surrogate policy gradient can be derived from Eq.~\ref{training objective}, written as:
\begin{align}
g(\theta; q)
&
 = \frac{1}{\sigma_{\mathrm{mix}}}\,\mathbb{E}\!\big[(r-\mu_r)\,S\big]
 = \frac{1}{\sigma_{\mathrm{mix}}}\,\mathrm{Cov}\!\big(r,\,S\big),
\end{align}
where 
$S = \nabla_\theta \log \pi_\theta(o\mid q)$ is the score function of the policy. 
Expanding the reward into components and rewriting each covariance term via its correlation coefficient, the gradient can be rewritten as: 
\begin{align}
\label{gradient}
g(\theta; q)
&= \frac{1}{\sigma_{\mathrm{mix}}}
   \sum_{j=1}^{N} \mathrm{Cov}(r^{(j)},S)
= \frac{\sigma_S}{\sigma_{\mathrm{mix}}}
   \sum_{j=1}^{N} \rho^{(j)}\,\sigma^{(j)}
\end{align}
where
\begin{align}
    \mathrm{Cov}(r^{(j)},S) = \rho^{(j)}\,\sigma^{(j)}\,\sigma_S
\end{align}
$\sigma_S$ and $\sigma^{(j)}$ are the standard deviations of $S$ and $r^{(j)}$, 
$\rho^{(j)}$ denotes the Pearson correlation coefficient between $r^{(j)}$ and $S$.  
Since $\sigma_S$ and $\sigma_{\mathrm{mix}}$ are common to all dimensions and simply rescale the gradient, we focus on $\rho^{(j)} \sigma^{(j)}$—and, for components with comparable $\rho^{(j)}$, the variance term $\sigma^{(j)}$ alone determines their relative weight. 
Thus $\sigma^{(j)}$ biases the effective weighting of each reward dimension. \textit{In
particular, high-variance components $\sigma^{(j)}$ induce larger covariance with $S$ 
and therefore dominate the mixed gradient, while low-variance components
are relatively suppressed.}

\noindent

\section{Method}
To overcome the above limitations, we propose Dr.~Seg, a GRPO-based method to enhance VLLMs for visual perception tasks. It comprises two key components: a Look-to-Confirm  strategy to enlarge the output space and a Distribution-Ranked Reward mechanism for fine-grained, stable rewards. The integration of the two components enables mutual enhancement  of richer exploration path and stabler reward feedback, leading to much higher performance.


\begin{figure*}[!h]
    \centering
    \includegraphics[width=0.8\linewidth]{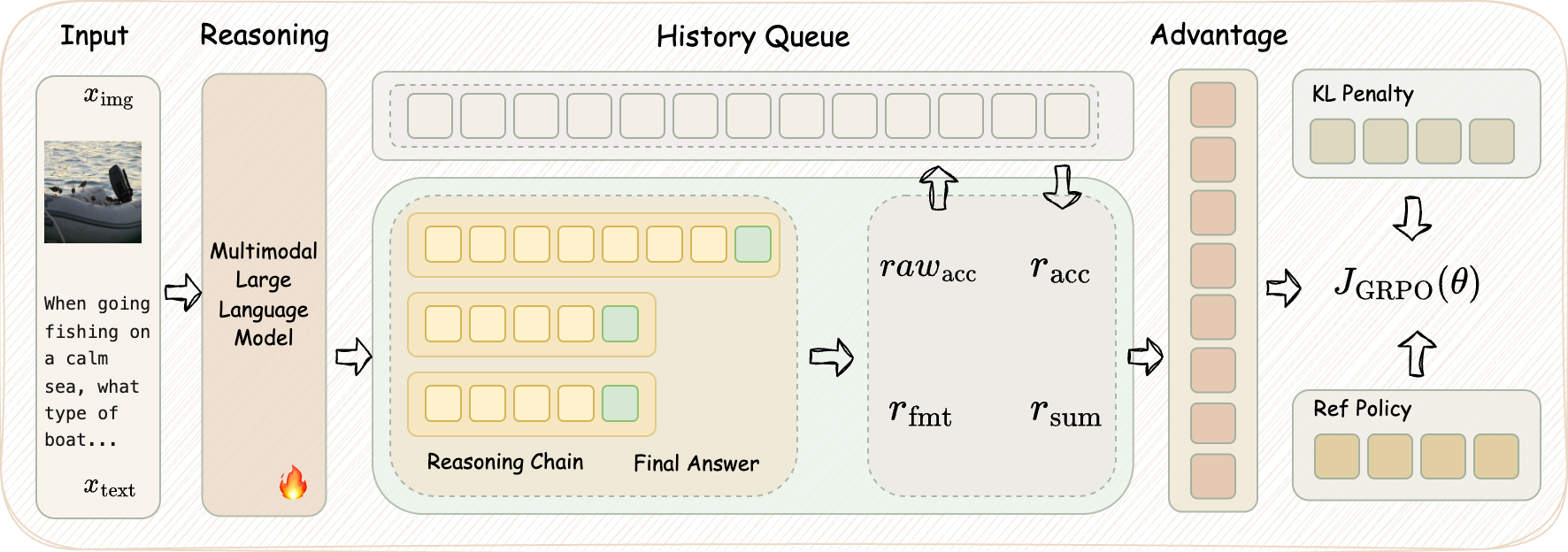}
    \caption{Overall training pipeline of Dr.~Seg.}
    \label{training pipeline}
\end{figure*}

\subsection{Architecture}
Following VisionReasoner~\cite{liu2025visionreasoner}, Dr.~Seg decouples the reasoning process from the segmentation process. 
We only train the VLLM, which is prompted to generate a set of boxes and points for each target object. 
These predicted boxes and points are then used to compute rewards and provide feedback during training under the GRPO~\cite{shao2024deepseekmath} framework. 
At evaluation time, the boxes and points produced by the VLLM are used as prompts to drive 
SAM2~\cite{ravi2024sam} for segmentation.

\subsection{Look-to-Confirm Visual Exploration}

\label{format reward}
\begin{figure}[h]
    \centering
    \includegraphics[width=1.0\linewidth]{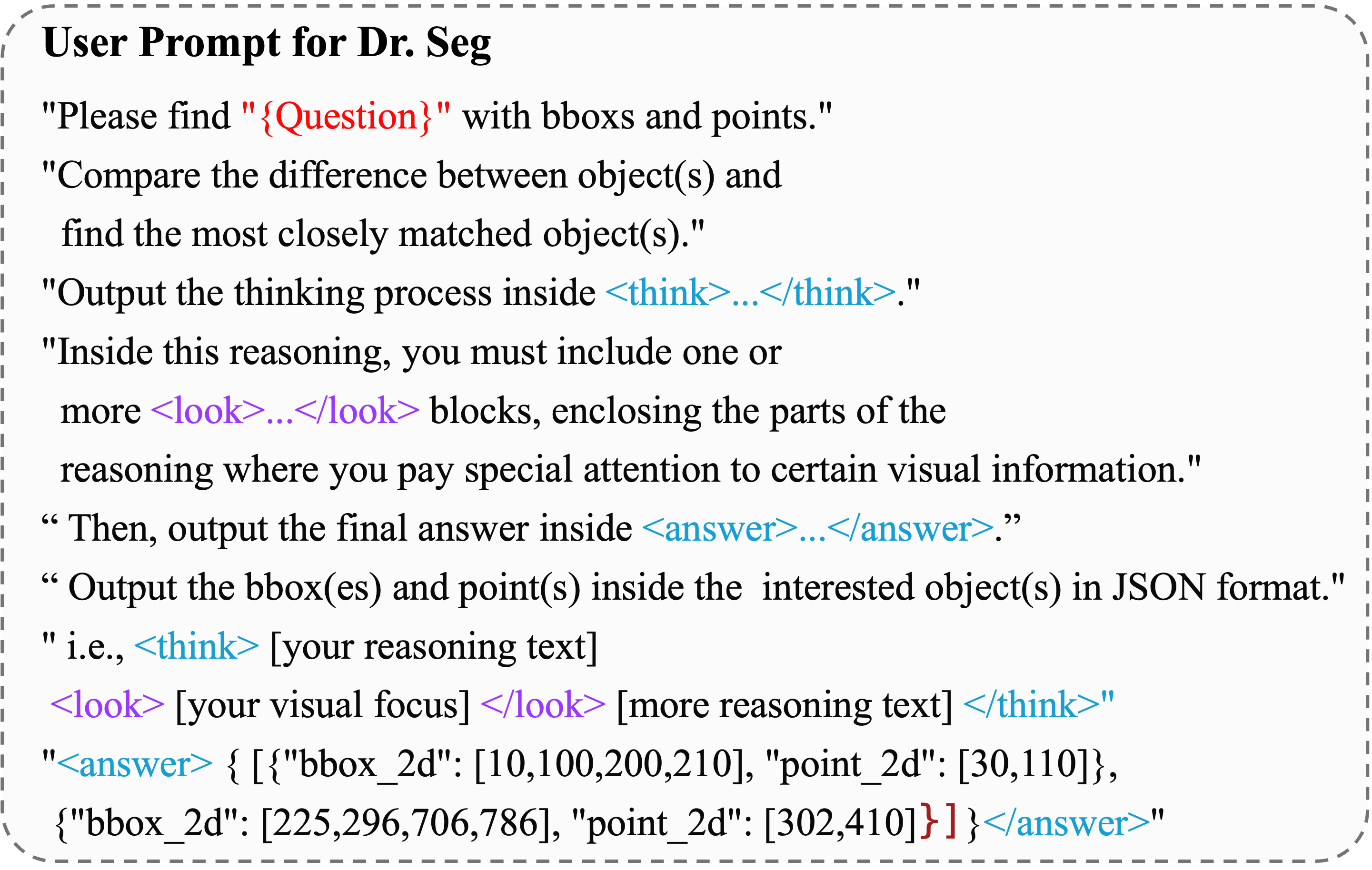}
    \caption{User prompt for Dr.~Seg.    
    }
    \label{fig:prompt}
    \vspace{-14pt}
\end{figure}
To encourage breadth exploration, we incorporate a visual exploration process into the 
reasoning path. Concretely, we prompt the model to explicitly mark salient visual evidence with a \textless look\textgreater~ tag. As presented in Fig.~\ref{fig:prompt}, the model is prompted to use \textless look\textgreater
~tags to explicitly enclose the parts of the reasoning that require special attention to visual information, and we assign a corresponding format reward to encourage adherence to this structure. 
This way, the model is forced to show its focus on certain visual information during reasoning, thus it looks for plausible visual cues from diverse perspectives with its pre-trained visual knowledge, e.g., looking at evidence such as shape specifics, material, spatial relations, thereby obtaining stronger generalization. 
This is inspired by Yang et al.~\cite{yang2025look}, who study the effect of explicitly injecting a reflection and verification process into model reasoning by inserting a \textless back\textgreater~ phase. Unlike their approach, we take the perspective of expanding the output search space, only asking the model to look over the image space, without performing any additional verification of its answers.


In addition, we follow prior work~\cite{liu2025seg,liu2025visionreasoner} and incorporate three additional format-related rewards and obtain the total format reward:
{
\begin{equation}
r_{\mathrm{fmt}}(o_i) = 
r_{\mathrm{look}}(o_i) + 
r_{\mathrm{think}}(o_i) +
r_{\mathrm{ans}}(o_i) +
r_{\mathrm{nr}}(o_i)
\label{eq:format_reward}
\end{equation} 
}
\noindent
Specifically, $r_{\text{look}}$ is 1.0 if the reasoning trace includes \textless look\textgreater....\textless /look\textgreater.
$r_{\text{think}}$ is 1.0 if the model outputs a thinking trace enclosed by
\textless think\textgreater~...\textless/think\textgreater~
and places the final answer within
\textless answer\textgreater~...\textless/answer\textgreater.
$r_{\text{ans}}$ is 1.0 if the final answer conforms to the restricted JSON format with correct key--value pairs.
$r_{\text{nr}}$ is 1.0 if the thinking trace is unique and non-repetitive.


\subsection{Distribution-Ranked Accuracy Reward}
\label{accuracy reward}

\noindent
\textbf{Overview and Quantile Mapping.}
To address the high-variance dominance problem, we develop a quantile mapping strategy. We maintain a fixed-length FIFO queue that stores recent accuracy vectors to approximate the empirical distribution of each metric. When computing the reward for a new output, each raw metric value is mapped to its empirical quantile (rank) within the queue, yielding a scale-invariant score. 
Specifically, given a candidate output $o_i$, we first define an $N$-dimensional raw accuracy vector, where each dimension corresponds to a perception-oriented evaluation metric:

\begin{equation}
\boldsymbol{x}(o_i) = (x_1, x_2, \dots, x_N) \in [0,1]^N
\end{equation}

\noindent
To obtain a scale-invariant and rank-based score comparable across batches, each component $x_j$ is mapped to its empirical quantile $q_j$ with respect to a FIFO rolling history queue $\mathcal{S}_t^{(j)}=(s_1^{(j)}, \dots, s_M^{(j)})$ maintained at training step $t$:
\begin{align}
q_j &= \frac{1}{M}\sum_{m=1}^{M} \mathbf{1}\!\big( s_m^{(j)} \leq x_j \big), \\
\boldsymbol{q}(o_i) &= (q_1,\dots,q_N)\in[0,1]^N,
\end{align}

\noindent
where $\mathbf{1}(\cdot)$ is the indicator function. By constructing this per-dimension Empirical Cumulative Distribution Function(ECDF)~\cite{dekking2005modern}, the mapping $T:\boldsymbol{x}\mapsto\boldsymbol{q}$ is coordinatewise monotone, bounded, and time-varying.



\noindent
\textbf{Final Reward Aggregation.}
The final distribution-ranked accuracy reward $r_{\mathrm{acc}}$ is aggregated across the $N$ quantile scores:

\begin{equation}
r_{\mathrm{acc}}(o_i) \;=\; \frac{1}{N} \sum_{j=1}^N q_j,
\end{equation}

\noindent
yielding a robust, time-adaptive scalar reward that reflects the relative standing of $o_i$ under the evolving performance distribution. Importantly, our FIFO–queue–based normalization maps each raw accuracy component $x_j$ to an empirical quantile $q_j \in [0,1]$, i.e., to its rank within the recent history of that metric. This rank-based, history-relative scale decouples each component’s effective gradient scale from its raw numerical magnitude: the difficulty of a sample is determined by its position in the evolving performance distribution rather than by the absolute value, as validated by our experiments.

\noindent
\textbf{Specialization ($N=3$).}
To obtain fine-grained reward, in our VLLM setting, the raw accuracy vector $\boldsymbol{x}$ comprises three essential perception metrics:
\noindent $x_1$ measures the alignment between the predicted bounding boxes $b_{\text{pred}}$ and ground-truth bounding boxes $b_{\text{gt}}$ (IoU).
\begin{equation}
x_1 = \mathrm{IoU}(b_{\text{pred}}, b_{\text{gt})}
\label{eq:x1}
\end{equation}

\noindent $x_2$ evaluates the prediction coverage consistency based on the counts of predicted $N_{\text{pre}}$ and ground-truth objects $N_{\text{gt}}$.

\begin{equation}
x_2 = \frac{\min(N_{\text{pre}}, N_{\text{gt}})}{\max(N_{\text{pre}}, N_{\text{gt}})},
\label{eq:x2}
\end{equation}

\noindent
$x_3$ quantifies the similarity between the predicted points $p_{\text{pred}}$ and ground-truth points $p_{\text{gt}}$. 
Let $d_{\text{pt}} = \lVert p_{\text{pred}} - p_{\text{gt}} \rVert_2$ denote their Euclidean distance; 
we then define
\begin{equation}
x_3 = g(d_{pt}),
\label{eq:x3}
\end{equation}
\begin{equation}
g(d_{pt}) =
\begin{cases}
1, & d_{pt} \leq \tau_{\min}, \\
\dfrac{\tau_{\max} - d_{pt}}{\tau_{\max} - \tau_{\min}}, & \tau_{\min} < d_{pt} < \tau_{\max}, \\
0, & d_{pt} \geq \tau_{\max}.
\end{cases}
\label{eq:g-def}
\end{equation}

\noindent
$\tau_{\min}$ and $\tau_{\max}$ define the bounds of the soft penalty region for the point guidance signal.

\begin{table*}[!h]  
\centering
\caption{Performance on in-distribution (ID) and out-of-distribution (OOD) benchmarks. $\dagger$indicates that the code was not open-sourced as of the submission deadline. $*$denotes results obtained from our independent experiments under the same experimental conditions. RCO, RCO+, and RCOg refer to RefCOCO, RefCOCO+, and RefCOCOg, respectively. - denotes not available.}
\label{tab:main results}
\setlength{\tabcolsep}{6pt}
\renewcommand{\arraystretch}{0.95}
\begin{tabular}{l|ccc|c|ccc|c|c}
\toprule
\multirow{4}{*}{\textbf{Method}} &
\multicolumn{4}{c|}{\textbf{In Distribution}} &
\multicolumn{4}{c|}{\textbf{Out of Distribution}} &
\textbf{Total}\\
\cmidrule(lr){2-5}\cmidrule(lr){6-9}\cmidrule(lr){10-10}
& \textbf{RCO} & \textbf{RCO+} & \textbf{RCOg} & \multirow{2}{*}{\textbf{ID avg\ }} 
& \multicolumn{2}{c}{\textbf{\ ReasonSeg\ \ \ }}

& \textbf{COCONut}
& \multirow{2}{*}{\textbf{OOD avg}}
& \multirow{2}{*} {\textbf{avg}}\\
\cmidrule(lr){2-4}\cmidrule(lr){6-8}
& \textit{testA} & \textit{testA} & \textit{test} & 
& \textit{\ val\ }
& \textit{test}
& \textit{val} & \\
\midrule
\multicolumn{10}{l}{\textcolor{gray}{Task-specific Models}} \\
LAVT~\cite{yang2022lavt} & 75.8 & 68.4 & 62.1 & 68.8 & - & - & - & -  \\
ReLA~\cite{liu2023gres}  & 76.5 & 71.0 & 66.5 & 71.3 & 22.4 & 21.3 & - & -  \\
\midrule
\multicolumn{10}{l}{\textcolor{gray}{Base 7B Models}} \\
Qwen2-VL~\cite{wang2024qwen2}    & 68.7 & 65.7 & 63.5 & 66.0 & 44.5 & 38.7 & 9.8*  & 31.0 & 46.9 \\
Qwen2.5-VL~\cite{bai2025qwen2}  & 79.9 & \textbf{76.8} & 72.8 & 76.5  & 56.9 & 52.1 & 31.8*  & 45.5 & 61.7 \\
\midrule
\multicolumn{10}{l}{\textcolor{gray}{SFT-based 7B VLLMs}} \\ 
LISA~\cite{lai2024lisa}        & 76.5 & 67.4  & 68.5 & 70.8 & 44.4 & 36.8 & - & - & -\\
GLaMM~\cite{rasheed2024glamm}      & 58.1  & 47.1  & 55.6  & 53.6 & - & -  & - & - & -\\
PixelLM~\cite{ren2024pixellm}     & 76.5 & 71.7 & 70.5 & 72.9 & - & - & - & - & - \\
GSVA~\cite{xia2024gsva}    & 78.9 & 69.6 & 73.3 & 73.9 & - & - & - & - & - \\
OMG-LLaVa~\cite{zhang2024omg}     & 79.8 & 73.0 & 71.9 & 74.9 & - & - & - & - & - \\
\midrule
\multicolumn{9}{l}{\textcolor{gray}{RL-based 7B VLLMs}}\\
Seg-Zero~\cite{liu2025seg}   & \textbf{80.3}  & 76.2 & 72.6 & 76.4 & 62.6  & 57.5 & 69.3* & 63.1 & 69.8 \\
Pixel-Think$\dagger$~\cite{wang2025pixelthink}    & 79.3  & 74.8 & 73.9 & 76.0 & 63.8  & 60.2 & - & - & - \\
VisionReasoner~\cite{liu2025visionreasoner}  & 78.9 & 74.9 & 71.3 & 75.0 & 66.3 & 63.6 & 78.1* & 69.3 & 72.2 \\
VisionReasoner* & ~~79.0* & ~~75.3* & ~~72.5* & 75.6 & ~~65.5* & ~~61.5* & 78.1*  & 68.4 & 72.0 \\
\textbf{Dr.~Seg (Ours)} & 80.2 & \textbf{76.8} & \textbf{74.2} & \textbf{77.1} & \textbf{67.8} & \textbf{65.6} & \textbf{79.6~ } & \textbf{71.0} & \textbf{74.0} \\
\bottomrule
\end{tabular}
\vspace{-6pt}
\end{table*}

\section{Experiments}

\subsection{Experiment Setup}
\textbf{Datasets and Evaluation Benchmarks.} 
Following~\cite{liu2025visionreasoner}, we adopt the same datasets and preparation strategy. 
The training dataset \textit{VisionReasoner\_multi\_object\_7k\_840} is constructed from four sources: LVIS~\cite{gupta2019lvis}, RefCOCOg~\cite{yu2016modeling}, gRefCOCO~\cite{liu2023gres}, and LISA++~\cite{yang2023lisa++}, 
resulting in approximately 7,000 training samples in total; refer to supplementary materials for more details. For evaluation, we follow prior work~\cite{liu2025seg,liu2025visionreasoner,wang2025pixelthink} and report gIoU, 
defined as the mean IoU across all segmentation masks. 
We adopt the same data splits as~\cite{liu2025visionreasoner,liu2025seg,wang2025pixelthink} across RefCOCO, RefCOCO+, RefCOCOg, and ReasonSeg, totaling approximately 10{,}000 samples for a fair comparison. The first three datasets serve as in-distribution benchmarks for referring expression segmentation, whereas ReasonSeg serves as the out-of-distribution benchmark for reasoning segmentation.

We additionally construct an evaluation setting to further assess multi-object segmentation. Specifically, we sample 665 images from the COCONut panoptic segmentation dataset~\cite{deng2025coconut}, selecting scenes that require segmenting multiple objects simultaneously. The resulting split contains an average of 5.14 target object instances per image and spans 68 categories, all within the original COCO taxonomy. Further details of the construction protocol are provided in the supplementary material.

\noindent\textbf{Running Quantile Estimator.}
For the distribution-ranked accuracy reward, we employ a quantile service that maintains a fixed-length FIFO queue of size 2,048, corresponding to the historical accuracy records from the past 16 training steps.
We initialize the queue with a full step of zeros before inserting actual records.
In practice, at each rollout, raw accuracy vectors from candidate outputs are first pushed into a temporary buffer, which is then flushed into the global queue at the end of each training step.

\noindent
\textbf{Implementation Details. }We train Dr.~Seg under the VERL~\cite{sheng2024hybridflow} framework, employing Qwen2.5-VL-7B~\cite{bai2025qwen2} and SAM2-Large~\cite{ravi2024sam} as the reasoning and segmentation models, respectively. Dr.~Seg is initialized with settings similar to VisionReasoner~\cite{liu2025visionreasoner}. We use a batch size of 16 and a learning rate of $1 \times 10^{-6}$. All training is conducted on 4 NVIDIA H800 GPUs for around 500 steps. For the distance-based reward mapping function $g(\cdot)$, we set the lower threshold to 30 and the upper threshold to 200.

\subsection{Main Results}
We evaluate Dr.~Seg against state-of-the-art VLLMs, including LISA ~\cite{lai2024lisa}, GLaMM~\cite{rasheed2024glamm}, PixelLM~\cite{ren2024pixellm}, GSVA~\cite{xia2024gsva}, OMG-LLaVa~\cite{zhang2024omg}, Seg-Zero~\cite{liu2025seg}, VisionReasoner~\cite{liu2025visionreasoner}, Pixel-Think~\cite{wang2025pixelthink}, Qwen2-VL~\cite{wang2024qwen2}, Qwen2.5-VL~\cite{bai2025qwen2}. We also compare task-specific models, including LAVT~\cite{yang2022lavt} and ReLA~\cite{liu2023gres}. For models that do not report gIoU, we use their cIoU reported in~\cite{liu2025visionreasoner} as an alternative. 

As shown in Tab.~\ref{tab:main results}, Dr.~Seg consistently achieves state-of-the-art performance across 5/6 benchmarks, outperforming other models in its class.
Notably, while prior studies~\cite{liu2025seg,liu2025visionreasoner} often attained state-of-the-art performance in either in-distribution or out-of-distribution environments alone, our model achieves new state-of-the-art results in both simultaneously.
We attribute this superior performance to the task-specific design tailored for perception-oriented settings, where our two proposed improvements work synergistically.
Refer to the supplementary material for qualitative results.

\subsection{Additional Task Results}
To further assess the effectiveness and generality of our approach, we additionally evaluate Dr.~Seg on object detection and counting under the same experimental settings as VisionReasoner~\cite{liu2025visionreasoner}. For detection, we use COCO~\cite{lin2014microsoft}. For counting, we use PixMoCount~\cite{deitke2025molmo} and CountBench~\cite{paiss2023teaching}.
As shown in Tab.~\ref{tab:detection and counting}, our method also achieves state-of-the-art.
\begin{table}[h]
  \caption{Performance on Detection and Counting tasks. We approximate COCO AP using the ratio of the bounding box area to the total image area to enable compatibility with COCOAPI.}
  \label{tab:detection and counting}
  \centering
  \renewcommand{\arraystretch}{0.95}
  \setlength{\tabcolsep}{6pt}
  \begin{tabular}{l|c|cc|c}
    \toprule
    \multirow{3}{*}{\textbf{Method}} &
    \multicolumn{1}{c|}{\textbf{Detection}} &
    \multicolumn{3}{c}{\textbf{Counting}}\\
    \cmidrule(lr){2-2}\cmidrule(lr){3-5}
    & \multicolumn{1}{c|}{\textbf{COCO}} &
      \multicolumn{2}{c|}{\textbf{Pixmo}} &
      \multicolumn{1}{c}{\textbf{Count}} \\
    \cmidrule(lr){2-2}\cmidrule(lr){3-4}\cmidrule(lr){5-5}
    & \multicolumn{1}{c|}{\textit{val}} &
      \textit{val} & \textit{test} & \textit{test} \\
    \midrule
    Qwen2-VL~\cite{wang2024qwen2}         & 28.3 & 29.9 & 48.0 & 76.5 \\
    Qwen2.5-VL~\cite{bai2025qwen2}       & 29.2 & 63.3 & 67.9 & 76.0 \\
    VisionReasoner~\cite{liu2025visionreasoner}  & 37.7 & 70.1 & 69.5 & 87.6 \\
    \textbf{Dr.\,Seg (Ours)} & \textbf{40.1} & \textbf{74.6} & \textbf{72.4} & \textbf{91.5} \\
    \bottomrule
  \end{tabular} 

  \vspace{-6pt}
\end{table}


\subsection{Ablation Study}
\label{ablation study}

\noindent We conduct ablation experiments to disentangle the contributions of different components in Dr.~Seg.

\noindent
\textbf{Effect of Look-to-Confirm  and Distribution-Ranked Reward.} 
We conduct an ablation study to disentangle the contributions of our two proposed components: (i) enforcing the \textless look\textgreater~format and (ii) adopting the distribution-ranked accuracy reward. Tab.~\ref{tab:ablation}  reports results under four settings. 

\begin{table}[h]
  \caption{Ablation study on Look-to-Confirm  and Distribution-Ranked Reward. LC and DR refer to Look-to-Confirm and Distribution-Ranked Reward, respectively.}
  \label{tab:ablation}
  \centering
  \begin{tabular}{ll|ccccc}
    \toprule
     LC & DR & RCO & RCO+ & RCOg & ReasonSeg\\
    \midrule
      &  & 79.0 & 75.3 & 72.5 & 65.5 \\
     \checkmark &  & 79.0 & 75.1 & 72.1 & 66.1 \\
      & \checkmark & 80.1 & 76.6 & 73.9 & 65.5 \\
     \checkmark & \checkmark & \textbf{80.2} & \textbf{76.8} & \textbf{74.2} & \textbf{67.8} \\
    \bottomrule
  \end{tabular}
\end{table}

The ablation results reveal the difference between the two modules. Look-to-Confirm encourages a broader output space during training, enhancing the model’s ability to generalize better on the out-of-distribution reasoning segmentation task, which requires more complex visual reasoning than training data, i.e., a 0.6 point improvement on ReasonSeg task. Yet due to the coarse binary reward, the learning of continuous predictions such as bounding boxes is affected, resulting in slightly reduced performance on the in-distribution accuracy. 
Distribution-Ranked Reward, on the other hand, improves in-distribution performance, i.e., 1.1, 1.5 and 1.8 absolute IoU gains on the RefCOCO, RefCOCO+, RefCOCOg, respectively. This indicates that it provides stable, fine-grained signals that improve the model's data fitting capability. However, since there is no regularization encouraging the output space exploration, the out-of-distribution ReasonSeg performance is not improved. Surprisingly, when the two mechanisms are combined, they work synergistically together, unlocking marked enhancement on both in-distribution and out-of-distribution tasks, demonstrating both the effectiveness of our design and the synergy between the two components.

\noindent\textbf{Effect of Normalization.}
To isolate the contribution of our distribution-ranked  normalization mechanism, we compare two settings under the same reward formulation:
(i) directly applying raw, unnormalized continuous rewards, and
(ii) applying distribution-ranked normalization through the FIFO queue.
As shown in Tab.~\ref{tab:queue}, the raw reward setting leads to a substantial degradation, which is consistent with our previous observations. We also present a snapshot of training rewards with rank-based normalization in Fig.~\ref{vis_bias}, showing its effectiveness. 
{
\begin{table}[h]
  \caption{Ablation on normalization strategy.}
  \label{tab:queue}
  \centering
  \begin{tabular}{lcccc}
    \toprule
    Normalization & RCO & RCO+ & RCOg & ReasonSeg \\
    \midrule
    Raw Reward & 78.7 & 75.0 & 71.5 & 64.4 \\
    Distribution-Ranked& \textbf{80.2} & \textbf{76.8} & \textbf{74.2} & \textbf{67.8} \\
    \bottomrule
  \end{tabular}
\end{table}
}

\begin{figure}[t]
    \centering
    \includegraphics[width=1\linewidth]{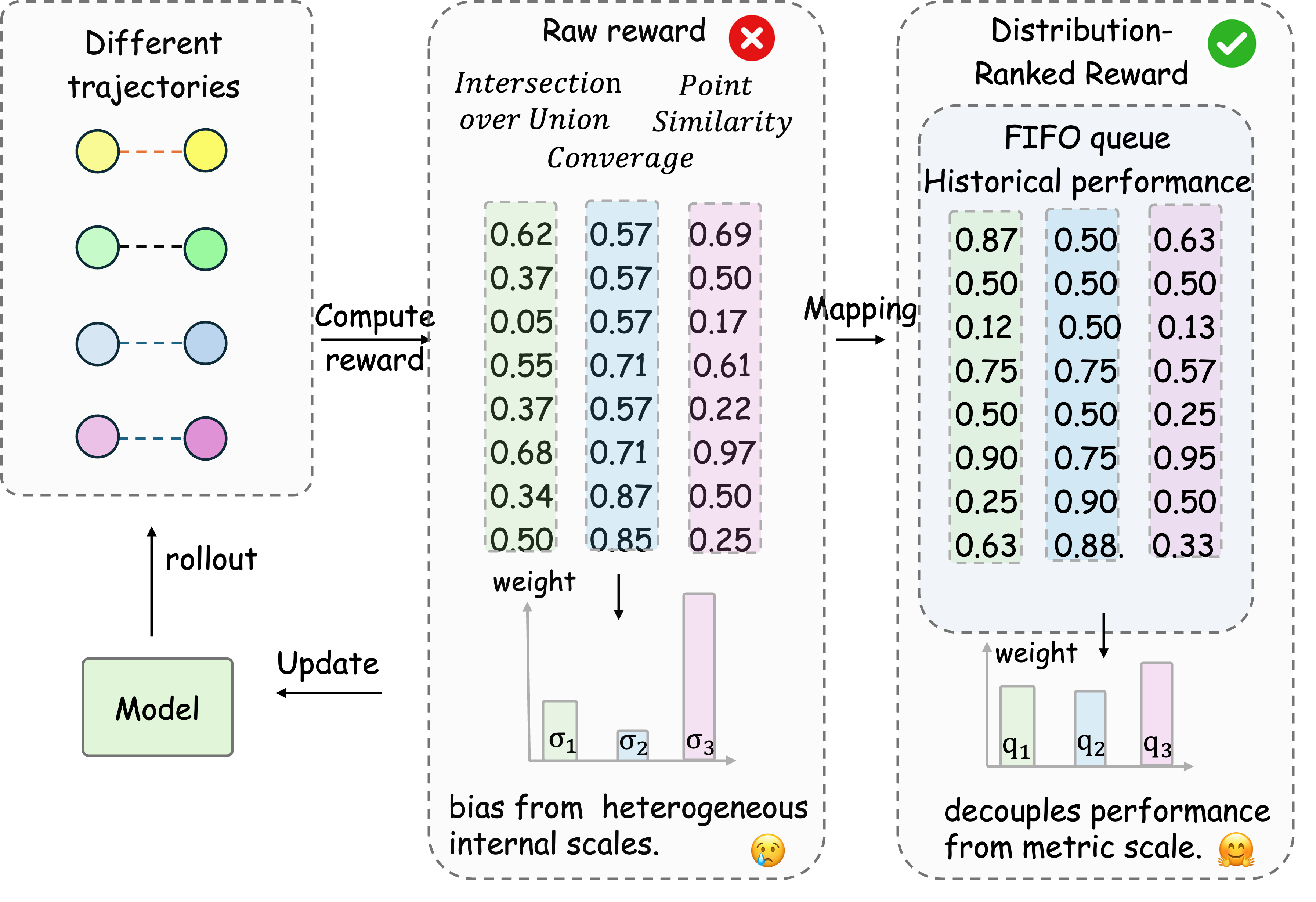}
    \caption{A snapshot of sample training rewards. Our Distribution-Ranked Reward maps each reward component to an empirical quantile $\mathbf{q} = [q_1, q_2, q_3] \in [0,1]^3$ using a FIFO history queue, obtaining rank-based score that is scale-invariant and overcoming high-variance domination.}
    \label{vis_bias}
    \vspace{-12pt}
\end{figure}

\noindent
\textbf{Effect of Queue Size.}
We vary the FIFO queue size for Distribution-Ranked Reward normalization. As shown in Tab.~\ref{tab:queuesize}, larger queues yield more stable and consistent gains on ID benchmarks. However, when the queue becomes too long, OOD performance starts to degrade—resembling the behavior of the raw-reward setting. We speculate that an overly long history queue makes the baseline overly stable: exploring new breadth directions naturally induces short-term performance fluctuations over several neighboring steps, and this over-stabilized design instead suppresses such exploration.

\begin{table}[h]
  \caption{Ablation on queue size (numbers in parentheses indicate past steps).}
  \label{tab:queuesize}
  \centering
  \begin{tabular}{lcccc}
    \toprule
    Queue Size & RCO & RCO+ & RCOg & ReasonSeg \\
    \midrule
    ~~128 (1) & 80.1 & 76.0 & 73.6 & 63.3 \\
    ~~512 (4) & 80.0 & 76.3 & 73.6 & 66.3 \\
    2048 (16) & 80.2 & 76.8 & 74.2 & \textbf{67.8} \\
    8192 (64) & \textbf{80.4} & \textbf{77.3} & \textbf{74.3} & 65.2 \\
    \bottomrule
  \end{tabular}
  \vspace{-10pt}
\end{table}

\section{Conclusion}
In this work, we revisit VLLM training for perception-oriented vision tasks and identify two overlooked aspects of current research: the entropy instability inherent to visual exploration and inappropriate reward design. 
Based on our analysis, we propose Dr.~Seg, a simple GRPO-based framework integrating a Look-to-Confirm  mechanism and a Distribution-Ranked Reward module. 
Without any architectural modifications, Dr.~Seg obtains state-of-the-art performance on a wide range of perception tasks such as referring segmentation and object detection. Meanwhile, it achieves strong generalization to out-of-distribution scenarios. 
Our work reveals the importance of perception-oriented design in both reasoning patterns and reliable rewards.

\noindent
\textbf{Limitations and Future Work. }
While effective, our study is limited to a subset of the broad vision tasks. Thus, how to transfer our method to other tasks (e.g., tracking, pose estimation, and video understanding) or non-visual modalities (e.g., point cloud) remains open. Additionally, the method only implicitly introduces the visual augmentation in the reasoning process of VLLM, which introduces no explicit feature enrichment from visual modality and may not be optimal for the dense object-level visual perception tasks.
Recent research has shown that explicit pathways such as spatially adaptive visual tokenization can effectively inject visual details and improve these core vision tasks. Thus synergistically combining both implicit and explicit visual injection remains future work.

\section*{Acknowledgment}

This work is supported by the National Science Foundation of China under Grant 62506249, the National Major Scientific Instruments and Equipments Development Project of National Natural Science Foundation of China under Grant 62427820, the Natural Science Foundation of Sichuan under grant 2024NSFSC1462, and the Fundamental Research Funds for the Central Universities under grant YJ202342.

{
    \small
    \bibliographystyle{ieeenat_fullname}
    \bibliography{main}
}

\clearpage
\setcounter{page}{1}
\maketitlesupplementary

\noindent
We provide supplementary material related to the main paper,
arranged as follows: 

\smallskip
\noindent1. More Experiment Details  and Ablations (Section~\ref{More Experiment Details})

\noindent2. Qualitative Results (Section~\ref{Qualitative Results})

\noindent3. Additional Results on REC (Section~\ref{Additional Results on REC})

\noindent4. Dataset Details (Section~\ref{Dataset Details})

\begin{figure*}[!t]
    \centering
    \includegraphics[width=1\linewidth]{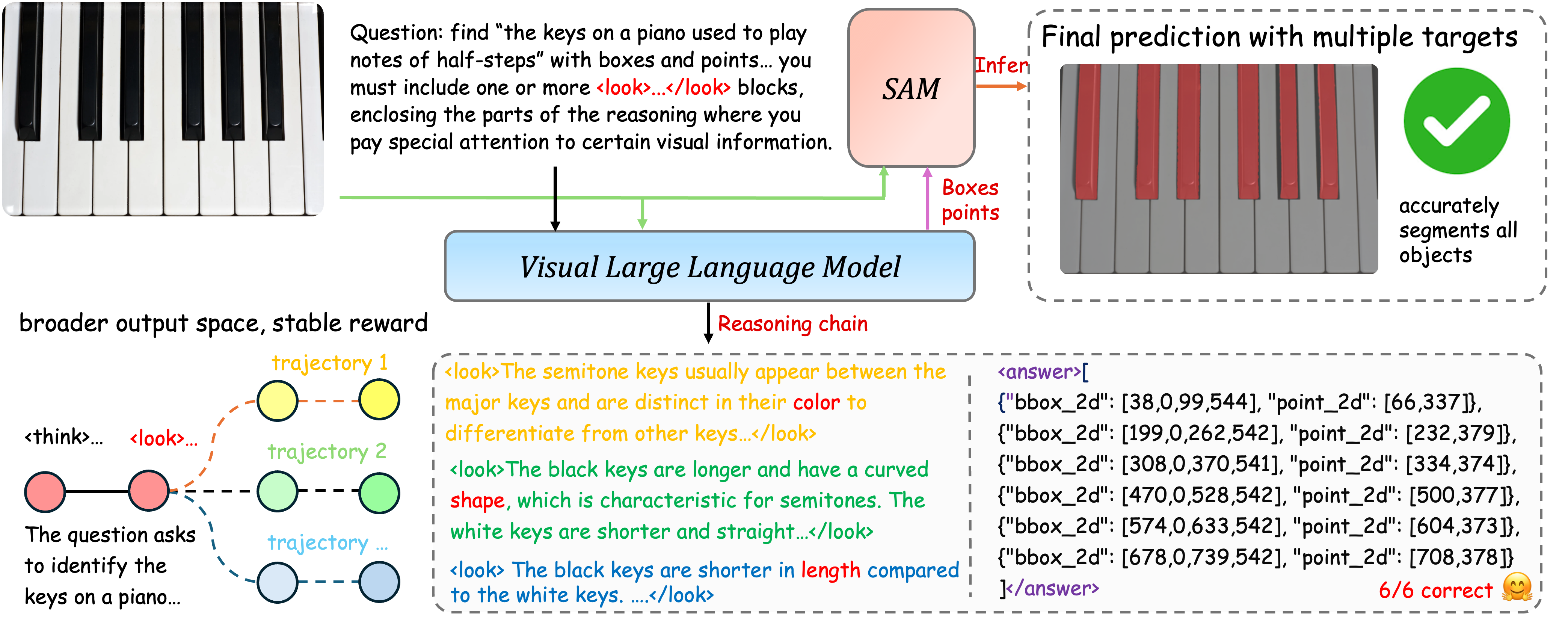}
    \caption{Supplementary illustration of the Dr.~Seg architecture. The model first performs a Look-to-Confirm exploration, generating multiple reasoning trajectories that describe diverse visual cues. During inference, these trajectories are used to prompt SAM to produce mask proposals. The example shown here demonstrates successful segmentation of all six semitone keys (6/6 correct).}
    \label{main_fig}
    \label{fig:placeholder}
\end{figure*}
    
\appendix

\section{More Experiment Details and Ablations}
\label{More Experiment Details}
\noindent\textbf{Details of Dr.~Seg. }
Our project is built on SAM~\cite{kirillov2023segment}, VERL~\cite{sheng2024hybridflow}, EasyR1~\cite{zheng2025easyr1}, and VisionReasoner~\cite{liu2025visionreasoner}. The global batch size is 16, with the micro batch size per device for updates and the micro batch size per device for experience collection both set to 2. The rollout number per sample is 8, and the KL-loss coefficient is \(1 \times 10^{-2}\).  
We use the AdamW optimizer with a learning rate of $1\times 10^{-6}$ and a weight decay of $1\times 10^{-2}$. 

\smallskip
\noindent
\textbf{Details of VisionReasoner with Raw Reward.}
In Sec.~3.3, we replace the binary rewards in VisionReasoner with unnormalized continuous rewards. 
Concretely, for the box IoU reward, we directly use the continuous IoU value without applying any threshold. 
For the box L1 reward and point L1 reward, we adopt the same linear scaling function in Eq.15. 
All other settings are kept the same as in the original VisionReasoner configuration.

\smallskip
\noindent
\textbf{COCO Precision/Recall Metrics.}
Following VisionReasoner~\cite{liu2025visionreasoner}, we compute mAP by assigning each predicted box a proxy score defined as the ratio of the predicted box area to the image area. We also evaluate Precision/Recall/F1 to avoid such approximation.
\begin{table}[h]
\vspace{-6pt}
\centering
\begin{tabular}{lcccccc}
\toprule
Method  & AP@0.5 & P@0.5 & R@0.5 & F1 \\
\midrule
Qwen2-VL    & 43.5  & 72.1 & 41.3 & 52.5 \\
Qwen2.5-VL  & 47.6 & 72.4 & 46.6 & 56.7 \\
VisionReasoner       & 57.3 & 62.6 & 63.9 & 63.3 \\
Dr.~Seg  & 59.0 & 69.6 & 64.3 & 66.8 \\
\bottomrule
\end{tabular}
\vspace{-3pt}
\caption{More detection metrics on COCO.}
\label{tab:coco_pr_metrics}
\vspace{-8pt}
\end{table}

\smallskip
\noindent
\textbf{Ablation on Response Length.}
To isolate whether the gains come from simply generating longer responses, we compare several test-time prompting variants that increase the average response length while keeping the underlying model unchanged. As shown in Tab.~\ref{tab:ablate_length}, merely elongating the output does not improve performance: both a generic Chain-of-Thought prompt and a Look-to-Confirm style prompt substantially increase the response length but degrade gIoU (e.g., 65.5 $\rightarrow$ 63.1 and 59.6). In contrast, Dr.~Seg achieves the best gIoU (67.8) with a similar length range, indicating that the improvement cannot be attributed to longer responses. Length denotes the word count, measured on the ReasonSeg val set.

\begin{table}[h]
\vspace{-2pt}
\centering
\begin{tabular}{@{}lcc@{}}
\toprule
Method & Length & gIoU\\
\midrule
Baseline & 77.8  & 65.5\\
+ Chain-of-Thought prompt & 117.0 & 63.1\\
+ Look-to-Confirm prompt  & 137.3 & 59.6\\
Dr.~Seg & 148.3 & 67.8\\
\bottomrule
\end{tabular}
\vspace{-3pt}
\caption{Ablation on response length using test-time prompts.}
\label{tab:ablate_length}
\vspace{-8pt}
\end{table}

\smallskip
\noindent
\textbf{Ablation on Format Reward.}
To further verify whether the gain could come from a particular output structure (e.g., the $\langle\text{look}\rangle$ tag) or learned semantic guidance, We conduct a controlled ablation that matches structure by retraining with the same format reward $R_{\text{format}}$, while changing the keyword inside the tag (e.g., $\langle\text{look}\rangle, \langle\text{wait}\rangle$) or using unrelated keywords. As shown in Tab.~\ref{tab:ablate_structure}, adding $R_{\text{format}}$ with random/wait keywords yields only marginal changes (64.9--65.6 gIoU), suggesting that format constraints alone are insufficient. Moreover, removing $R_{\text{format}}$ (Look$^{*}$) fails to stably generate the tag and degenerates toward the baseline, indicating $R_{\text{format}}$ is necessary for stable structured behavior.

\begin{table}[h]
\vspace{-2pt}
\centering
\setlength{\tabcolsep}{3.0pt}
\begin{tabular}{@{}lccc@{}}
\toprule
Keyword & $R_{\text{format}}$ & Length & gIoU\\
\midrule
Random & $\checkmark$ & 89.6  & 64.9\\
Wait   & $\checkmark$ & 95.3  & 65.6\\
Look$^{*}$ &  & 73.6 & 65.8\\
Look   & $\checkmark$ & 148.3 & 67.8\\
\bottomrule
\end{tabular}
\vspace{-3pt}
\caption{Ablation on response structure by retraining with/without the format reward $R_{\text{format}}$ and varying the tag keyword. Look$^{*}$: without $R_{\text{format}}$, the model fails to stably generate $\langle\text{look}\rangle$ and degrades toward the baseline.}
\label{tab:ablate_structure}
\vspace{-16pt}
\end{table}

\smallskip
\noindent
\textbf{Ablation on Accuracy Reward Components.}
To clarify the contribution of each accuracy component, we ablate the reward composition used in $R_{\text{accuracy}}$ by progressively adding $x_1$ (IoU), $x_2$ (count), and $x_3$ (point) in Tab.~\ref{tab:ablate_racc}. 

\begin{table}[h]
\vspace{-2pt}
\centering
\setlength{\tabcolsep}{3.2pt}
\begin{tabular}{@{}lcc@{}}
\toprule
Setting & RefCOCO testA & ReasonSeg val\\
\midrule
$w$ & 35.6 & 24.1\\
$w\,x_1$ & 72.9 & 57.8\\
$w\,x_1,x_2$ & 79.8 & 58.3\\
$w\,x_1,x_2,x_3$ & 80.2 & 67.8\\
\bottomrule
\end{tabular}
\vspace{-3pt}
\caption{Ablation on $R_{\text{accuracy}}$.}
\label{tab:ablate_racc}
\vspace{-8pt}
\end{table}

\section{Qualitative Results}
\label{Qualitative Results}
We provide a qualitative example in Fig.~\ref{main_fig} and more visualizations in Fig.~\ref{fig:dr_seg_example}. As illustrated in Fig.~\ref{main_fig}, prior methods~\cite{liu2025visionreasoner} struggle in complex multi-object segmentation scenarios, often miscounting target instances due to over- or under-segmentation. For example, in the piano half-step case, the baseline model correctly identifies only 4 out of 6 candidate keys.
We argue that these errors stem from two coupled factors:
(1) binary rewards fail to reflect fine-grained performance differences, providing only coarse supervision; and
(2) the restricted output space limits the model’s ability to explore diverse reasoning trajectories, causing incorrect object enumeration.
Our method alleviates these issues: the proposed Look-to-Confirm mechanism expands the model’s output space by encouraging exploration of alternative visual cues, while the Distribution-Ranked Reward delivers stable fine-grained feedback independent of metric scales.
Together, these components yield substantially more accurate object counts. We provide more visualizations in Fig.~\ref{fig:dr_seg_example}, highlighting the effectiveness of our method in challenging multi-object settings.

\section{Additional Results on REC}
\label{Additional Results on REC}
We also evaluate our models' performance on the referring expression comprehension (REC) task, with results reported in Tab.~\ref{REC}. 
Our model also achieves superior performance compared with previous methods~\cite{liu2025visionreasoner} and base models~\cite{bai2025qwen2,wang2024qwen2}. Specifically, it outperforms the baseline VisionReasoner on all three REC datasets, with an average improvement of 0.7 absolute points.

\section{Dataset Details}
\label{Dataset Details}
\noindent\textbf{Training Dataset Statistics. } We use the \textit{VisionReasoner\_multi\_object\_7k\_840} dataset~\cite{liu2025visionreasoner}, which was constructed by the authors of VisionReasoner from four sources: LVIS~\cite{gupta2019lvis}, RefCOCOg~\cite{yu2016modeling}, gRefCOCO~\cite{liu2023gres}, and LISA++~\cite{yang2023lisa++}. It is formed by sampling roughly 1,800 examples from the training split of each dataset, without any special data partitioning criteria, resulting in a balanced mixture across the four sources.

\smallskip
\noindent\textbf{Evaluation Dataset Statistics. }
Table~\ref{tab:detail dataset} summarizes the statistics of all evaluation datasets used in this study. For a fair comparison, we follow prior work~\cite{liu2025seg,liu2025visionreasoner} and adopt the same data splits as in their experiments; we report the number of evaluation samples. 
The benchmark suite covers (i) referring expression segmentation (RefCOCO, RefCOCO+, RefCOCOg~\cite{yu2016modeling}); (ii) reasoning-oriented segmentation (ReasonSeg~\cite{lai2024lisa}); (iii) our self-constructed multi-object split (COCONut; details in the next paragraph); (iv) object detection (MS~COCO~\cite{lin2014microsoft}); (v) object counting (Pixmo-Count~\cite{deitke2025molmo}, CountBench~\cite{paiss2023teaching}) and (vi) referring expression comprehension (RefCOCO, RefCOCO+, RefCOCOg).

\smallskip
\noindent\textbf{Evaluation Metrics. }
For object detection on COCO, we adopt the standard AP metric computed using the COCO API. For referring expression comprehension on RefCOCO series, we use bbox AP, which measures detection accuracy at an IoU threshold of 0.5. For object segmentation on RefCOCO series and ReasonSeg, we use gIoU, computed as the mean IoU across all segmentation masks. For counting tasks, we use count accuracy as the evaluation metric.
 
\begin{table}[h]
\centering
\caption{Statistics of evaluation benchmarks. Numbers combine validation and test splits where applicable. DET, REC, SEG, and CNT denote Detection, Referring Expression Comprehension, Referring Expression Segmentation, and Counting, respectively.}
\label{tab:detail dataset}
\begin{tabular}{l l r}
\toprule
\textbf{Dataset} & \textbf{Split} & \textbf{\# of samples} \\
\midrule
\multirow{1}{*}{DET}
  & COCO        & 36,781 \\
\midrule
\multirow{3}{*}{REC}
  & RefCOCO     & 5,786 \\
  & RefCOCO+    & 5,060 \\
  & RefCOCOg    & 7,596 \\
\midrule
\multirow{5}{*}{SEG}
  & RefCOCO     & 1,975 \\
  & RefCOCO+    & 1,975 \\
  & RefCOCOg    & 5,023 \\
  & ReasonSeg   & 979 \\
  & COCONut     & 665 \\
\midrule
\multirow{2}{*}{CNT}
  & Pixmo-Count & 1,064 \\
  & CountBench  & 504 \\
\midrule
\multicolumn{2}{l}{\textbf{SUM}} & \textbf{67,408} \\
\bottomrule
\end{tabular}
\end{table}

\begin{table*}[]  
\centering
\caption{Performance on the referring expression comprehension task.}
\label{REC}
\setlength{\tabcolsep}{6pt}
\renewcommand{\arraystretch}{0.95}
\begin{tabular}{l|cccccc|c}
\toprule
\multirow{4}{*}{\textbf{Method}} &
\multicolumn{7}{c}{\textbf{Referring Expression Comprehension }}\\
\cmidrule(lr){2-8}
& \multicolumn{2}{c}{\textbf{\ RefCOCO\ \ \ }}
& \multicolumn{2}{c}{\textbf{\ RefCOCO+\ \ \ }}
& \multicolumn{2}{c|}{\textbf{\ RefCOCOg\ \ \ }}
& \multirow{2}{*} {\textbf{avg}}\\
\cmidrule(lr){2-7}
& \textit{val} & \textit{testA} & \textit{val} &  \textit{testA} & \textit{val} & \textit{test} & \\
\midrule
\multicolumn{8}{l}{\textcolor{gray}{Task-specific Models}} \\
DQ-DETR~\cite{liu2023dq} & 88.6& 91.0 &81.7 &86.2 &82.8 &83.4&85.6\\
Grounding-DINO-T(ft)~\cite{liu2024grounding} &89.2& 91.9 &81.1 &87.4 &84.2 &84.9 & 86.5\\
Grounding-DINO-L(ft)~\cite{liu2024grounding} & \textbf{90.6} & \textbf{93.2} & 82.8&  \textbf{89.0} & 86.1 & 87.0 & \textbf{88.1} \\
\midrule
\multicolumn{8}{l}{\textcolor{gray}{7B Visual Large Language Models}} \\
Qwen2-VL~\cite{wang2024qwen2}      & 80.8 & 83.9 & 72.5&  76.5 & 77.3 & 78.2 & 78.2 \\
Qwen2.5-VL~\cite{bai2025qwen2}    & 88.8 & 91.7 & 82.3 & 88.2 & 84.7 & 85.7 & 86.9 \\
Shikra~\cite{chen2023shikra}     & 87.0 & 90.6 & 81.6&  87.4 & 82.3 & 82.2 & 85.2 \\
Ferret-v2~\cite{zhang2024ferret}     & 87.5 & 91.3 & 80.8&  87.4 & 83.9 & 84.8 & 86.0 \\
InternVL2-8B~\cite{chen2024far} & 87.1 & 91.1 & 79.8&  87.9 & 82.7 & 82.7 & 85.2 \\

VisionReasoner~\cite{liu2025visionreasoner} &88.6 &90.6 &\textbf{83.6} &87.9 &86.1& 87.5 & 87.4\\
\textbf{Dr.~Seg (Ours)} &89.4 &91.5	&\textbf{83.6}	&88.0	&\textbf{88.1}	&\textbf{88.0}   &\textbf{88.1}\\
\bottomrule
\end{tabular}
\end{table*}

\smallskip
\noindent\textbf{Details of COCONut Benchmark.}
\begin{figure}[h]
    \centering
    \includegraphics[width=0.8\linewidth]{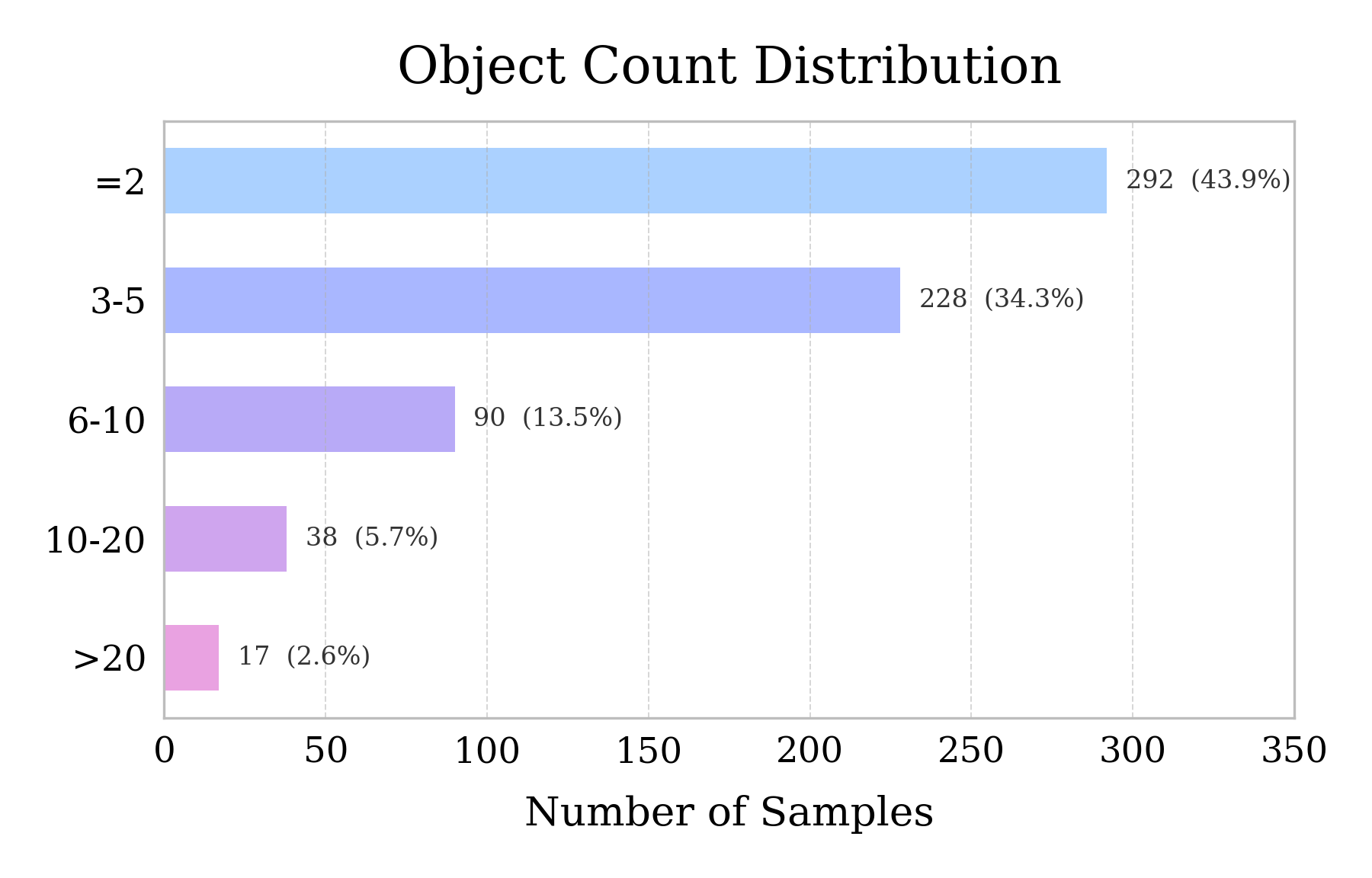}
    \caption{Object count distribution of the COCONut multi-object split.}
    \label{fig:dataset_dist}
\end{figure}
\begin{figure}[h]
    \centering
    \includegraphics[width=0.8\linewidth]{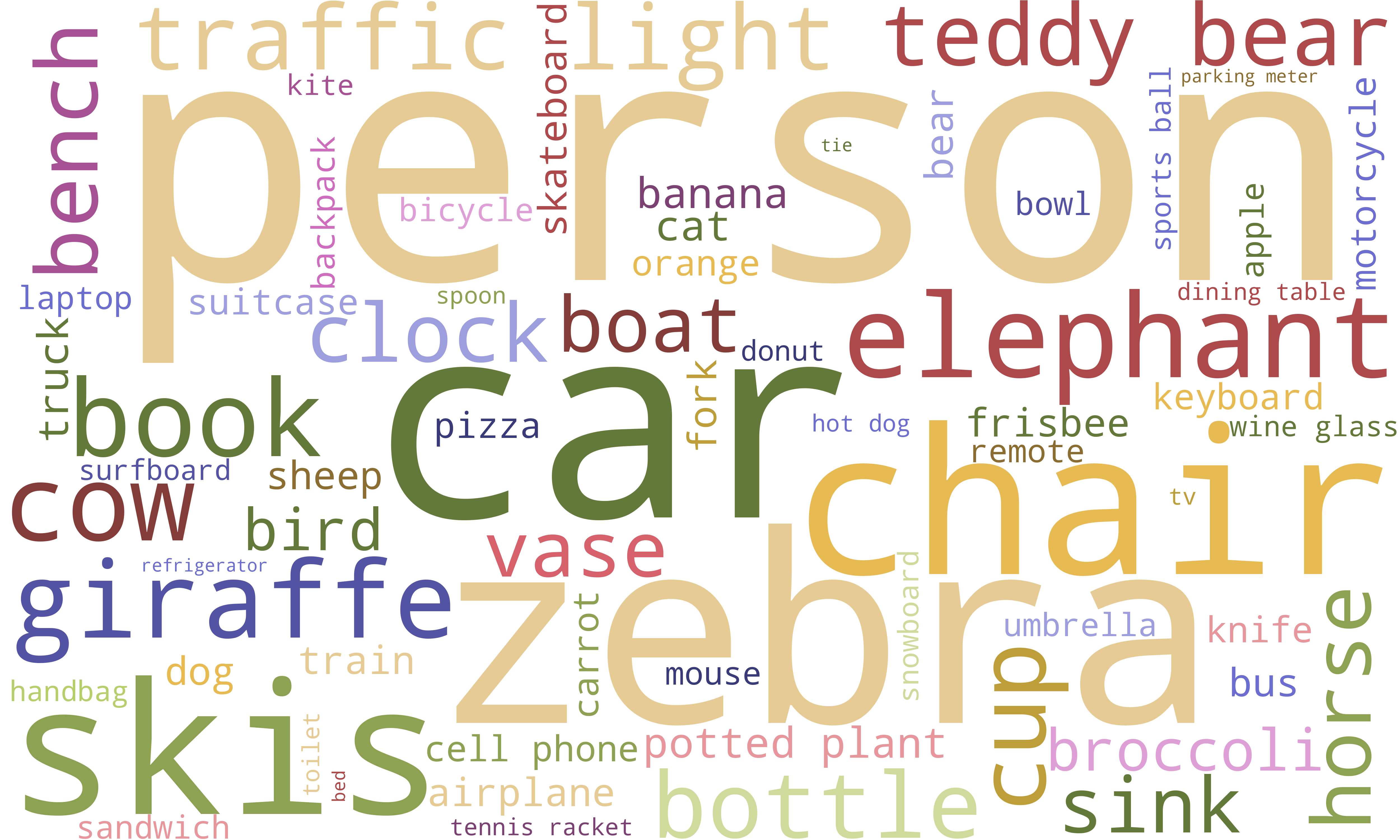}
    \caption{Categories included in our COCONut dataset, all of which are subclasses of the original COCO label set.}
    \label{fig:wordcloud}
\end{figure}
Since publicly available datasets for complex, multi-object segmentation remain scarce, we additionally construct a benchmark from COCONut val split~\cite{deng2025coconut}, a dataset that couples panoptic segmentation with grounded captions.

\smallskip
\noindent\textbf{Sampling Protocol.}
We sample 1,000 images from the official validation split and, after filtering, retain 665 images. The procedure is as follows:

\begin{enumerate}
\item \textbf{Candidate filtering}:
We discard images without annotations. For each remaining image, we count per-class instance frequencies and keep only those images in which at least one class appears at least twice. For each retained image, we collect the set of classes that satisfy this criterion and randomly choose one target class from this set.
\item \textbf{Union-mask construction}:
For the chosen class, we build a binary mask for every instance of that class and take the pixel-wise OR over all such masks to obtain the class-level union mask across the full image.
\item \textbf{Text prompt}:
For each sample, we form the instruction \texttt{All \textless category\textgreater}, indicating “segment all instances of the specified class.” 
\item \textbf{Quality Control}: 
Finally, we perform a light manual pass to remove low-quality cases (e.g., extremely small targets or ambiguous class references). The remainder constitutes our dataset.
\end{enumerate}
\begin{figure*}[h]
    \centering
    \includegraphics[width=0.95\linewidth]{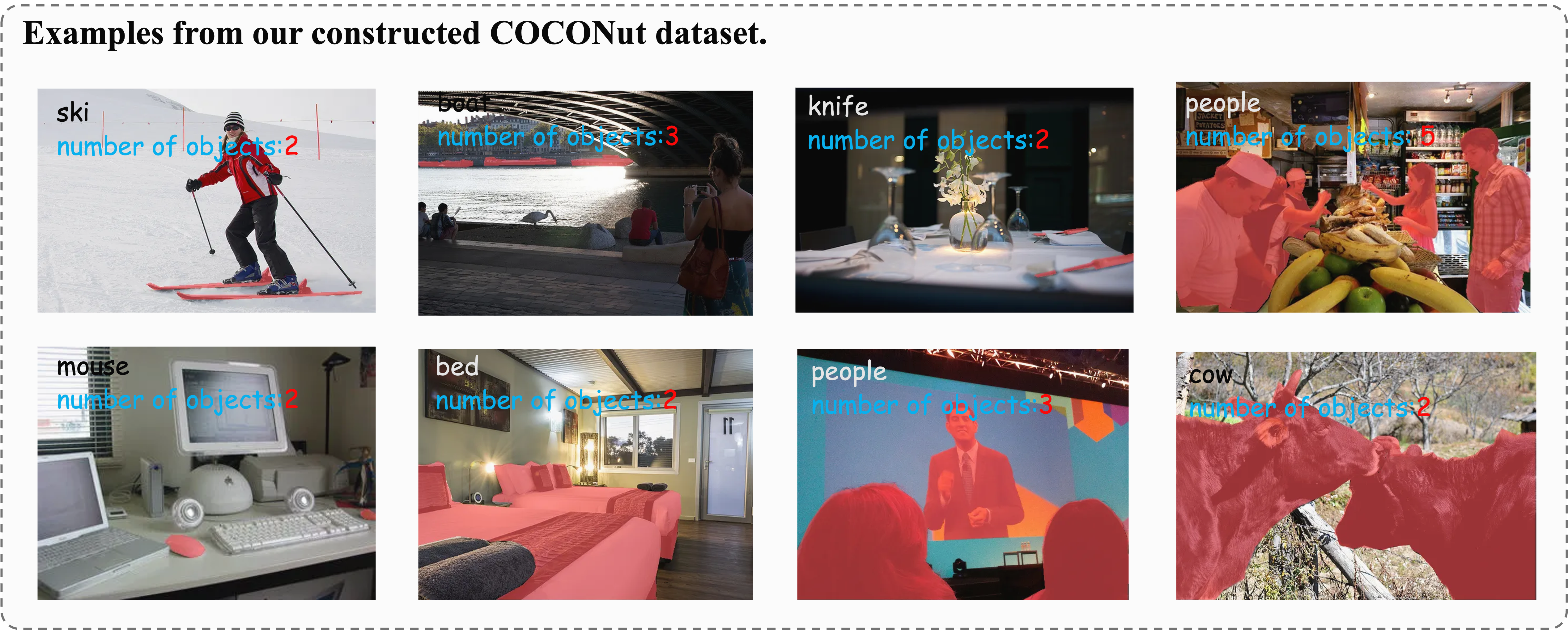}
    \caption{Visualization of our COCONut dataset, illustrating diverse object categories and a wide range of object counts per image.}\label{fig:dataset_example}
\end{figure*}
\smallskip
\noindent\textbf{Goal and Characteristics.}
This benchmark evaluates class-level multi-instance union reasoning: given a natural image and a concise class prompt, the model must output the combined region covering all instances of that class. Unlike typical single-instance prompts, this setting stresses coverage completeness over multiple same-class objects, which better reflects crowded or overlapping real-world scenes. 
We also provide summary statistics and qualitative examples in Fig.~\ref{fig:dataset_dist} and Fig.~\ref{fig:dataset_example}.

\begin{figure*}[h]
    \centering
    \includegraphics[width=0.95\linewidth]{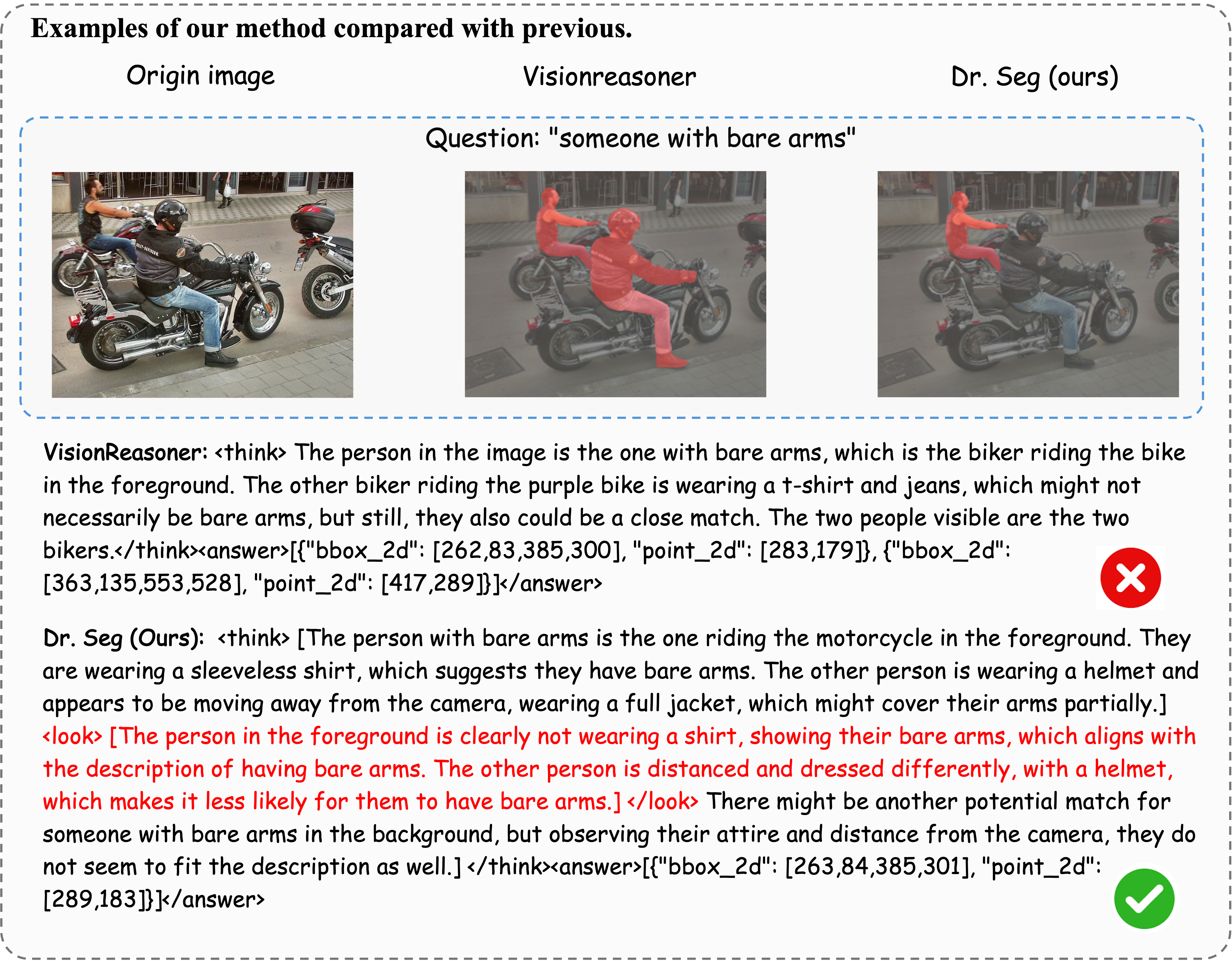}
    \caption{A representative failure case of prior methods. The model initially attends to the correct person but drifts away as the chain-of-thought unfolds, leading to over or under counting the target instances. We attribute this behavior to excessive noise in the reward design: collapsing raw rewards into binary signals discards information and introduces noise. This noise is then further amplified when heterogeneous metric scales are summed together.
}\label{fig:dr_seg_example}
\end{figure*}
\begin{figure*}[h]
    \centering
    \includegraphics[width=1.0\linewidth]{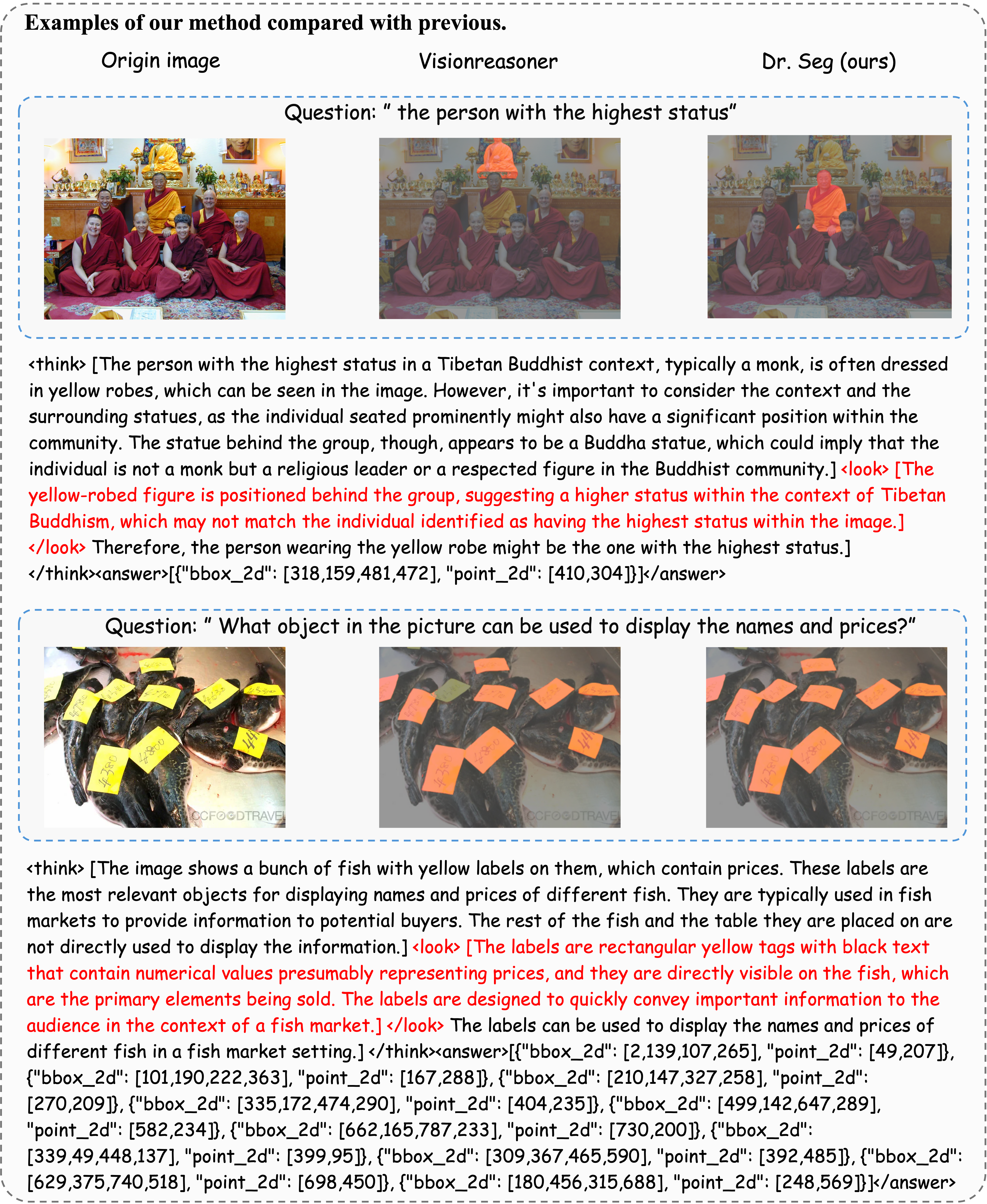}
    \caption{Visualization of the proposed Dr.~Seg compared with VisionReasoner.}\label{fig:dr_seg_example2}
\end{figure*}
\begin{figure*}[h]
    \centering
    \includegraphics[width=1.0\linewidth]{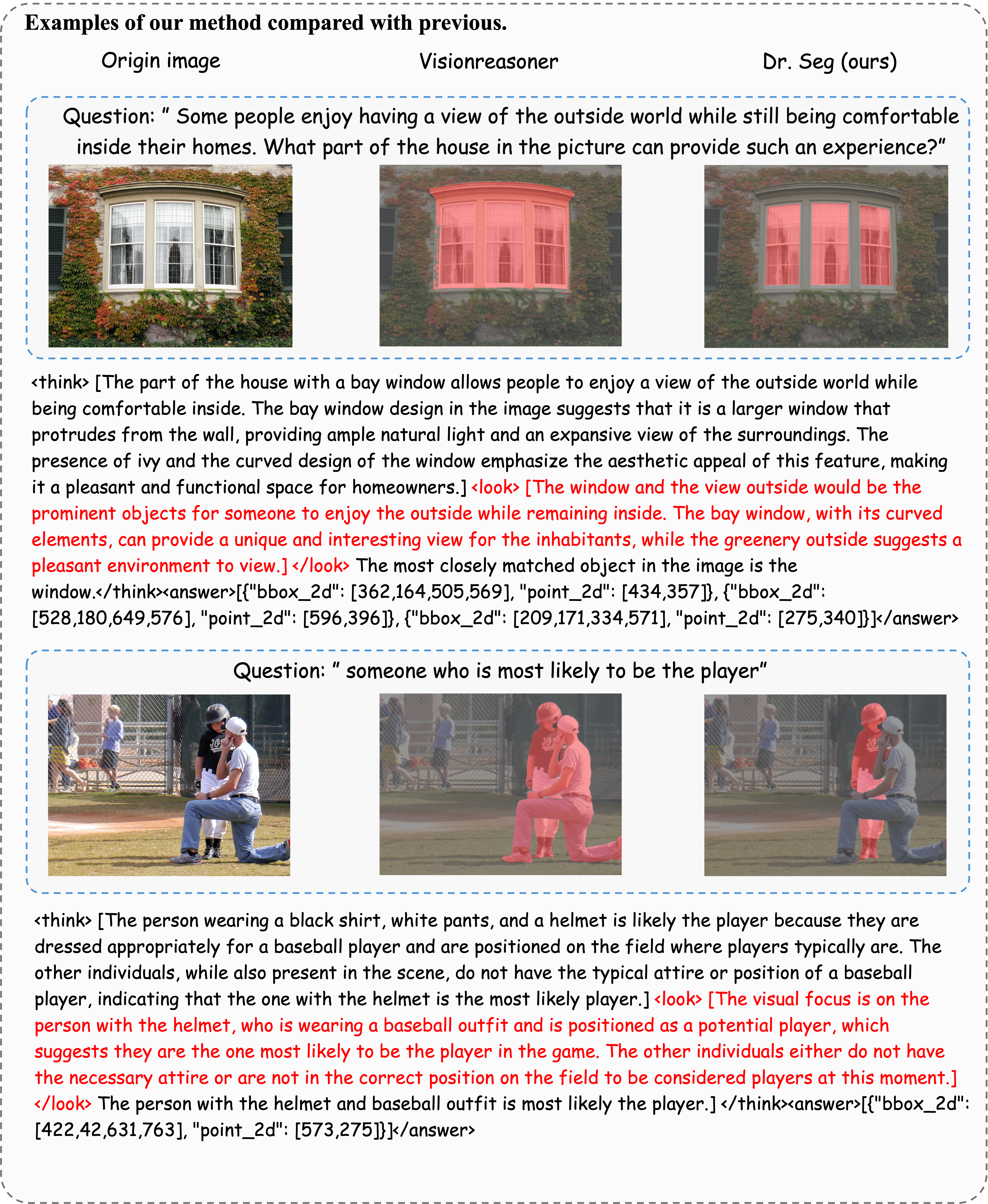}
    \caption{More visualization of the proposed Dr.~Seg compared with VisionReasoner.}\label{fig:dr_seg_example2}
\end{figure*}
\clearpage

\end{document}